\pdfoutput=1

\documentclass[11pt]{article}

\usepackage[]{EMNLP2023}

\usepackage{times}
\usepackage{latexsym}

\usepackage{bm}
\usepackage{xcolor}
\usepackage{multirow}
\usepackage{makecell}
\usepackage{textcomp}
\usepackage{graphicx}
\newcommand{\ie}{\textit{i}.\textit{e}., }
\newcommand{\eg}{\textit{e}.\textit{g}., }
\usepackage{amssymb}
\usepackage{amsmath}
\usepackage{comment}

\usepackage{newfloat}
\usepackage{caption}
\usepackage{tablefootnote}
\usepackage{booktabs}
\usepackage[T1]{fontenc}

\usepackage[utf8]{inputenc}

\usepackage{microtype}

\usepackage{inconsolata}

\title{HEAR: Hearing Enhanced Audio Response for Video-grounded Dialogue}

\author{\text{Sunjae Yoon}$^{\dagger}$ \hspace{0.5cm} \text{Dahyun Kim}$^{\dagger}$ \hspace{0.5cm} \text{Eunseop Yoon}$^{\dagger}$ \hspace{0.5cm} \text{Hee Suk Yoon}$^{\dagger}$  \\ {\bf Junyeong Kim}$^{\ddagger}$ \hspace{0.5cm} {\bf Chang D. Yoo}$^{\dagger}$\Thanks{Corresponding author} \\
         $^{\dagger}$Korea Advanced Institute of Science and Technology (KAIST) \\
         $^{\ddagger}$Chung-Ang University \\ \texttt{\{sunjae.yoon,dahyun.kim,esyoon97,hskyoon,cd\_yoo\}@kaist.ac.kr}\\ \texttt{junyeongkim@cau.ac.kr}}

\begin{document}
\maketitle
\begin{abstract}
Video-grounded Dialogue (VGD) aims to answer questions regarding a given multi-modal input comprising video, audio, and dialogue history. Although there have been numerous efforts in developing VGD systems to improve the quality of their responses, existing systems are competent only to incorporate the information in the video and text and tend to struggle in extracting the necessary information from the audio when generating appropriate responses to the question. The VGD system seems to be deaf, and thus, we coin this symptom of current systems' ignoring audio data as a deaf response. To overcome the deaf response problem, Hearing Enhanced Audio Response (HEAR) framework is proposed to perform sensible listening by selectively attending to audio whenever the question requires it. The HEAR framework enhances the accuracy and audibility of VGD systems in a model-agnostic manner. HEAR is validated on VGD datasets (i.e., AVSD@DSTC7 and AVSD@DSTC8) and shows effectiveness with various VGD systems. The project is available at: \href{https://github.com/dbstjswo505/HEAR}{\texttt{github.com/dbstjswo505/HEAR}}.
\end{abstract}
%
%
%
%
%
\section{Introduction}

\begin{figure}[t]
  \centering
  \includegraphics[width=\linewidth]{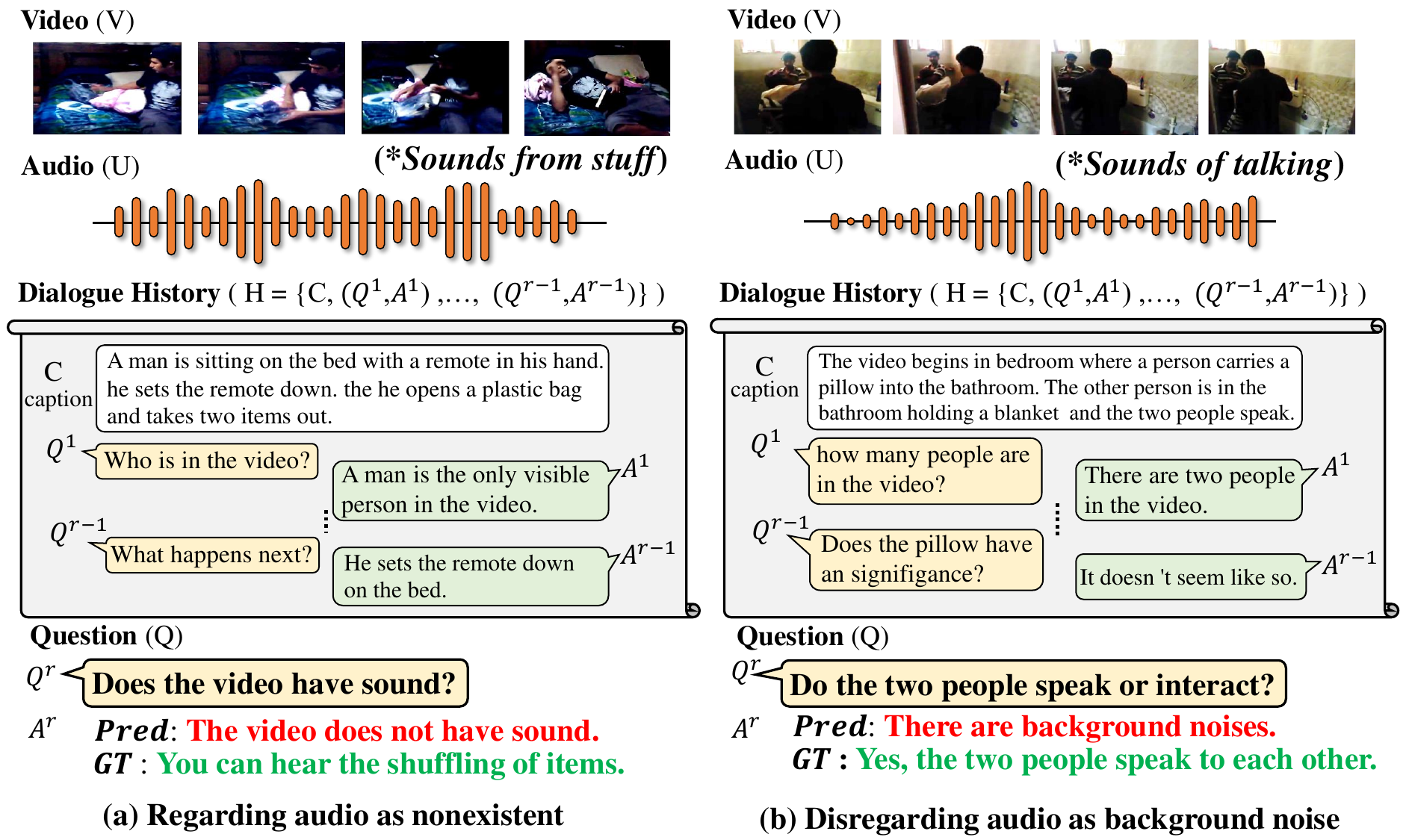}
  \caption{Current VGD system's deaf responses on questions about audio: (a) Audio is considered not present and (b) Audio is disregarded as background noise.}
  \label{fig:1}
\end{figure}

One of the desiderata in our vision-language community is to
build conversational agents that can look, listen, think and speak as humans.
These agents can potentially be deployed in various subsections of society, including education, security, entertainment, and visual or other impairments.
To promote natural conversation between humans and the conversational agents, the Video-grounded Dialogue (VGD) task \cite{alamri2019audio} has been designed, aiming to respond to the questions regarding a given multimodal input comprising video, audio, and dialogue history.
To be specific, as shown in Figure \ref{fig:1}, given video $V$, audio $U$, dialogue history composed of caption $C$ and past rounds of $Q\&A$ pairs $H = \{C, (Q^{1}, A^{1}), ...,(Q^{r-1}, A^{r-1})\}$, and current $r$-th round question $Q^{r}$, VGD system is expected to answer in free-form $A^{r}$ to the question $Q^{r}$.
For this VGD task, multi-modal interaction has been a popular solution, including transformer \cite{vaswani2017attention}, where many studies concerning modality interactions are performed to boost performance and improve the efficiency of VGD systems.
%

Unfortunately, these multi-modal interactions focus only on finding the joint representations between video and language. As a consequence, current agents tend to ignore audio in generating the response. This symptom of ignoring input audio in responding to the question will be referred to as the `deaf response'.
%
Figure \ref{fig:1} represents examples of deaf responses of existing VGD systems \cite{yoon-etal-2022-information,li2021bridging}. 
To the question ``Does the video have sound?'' in Figure \ref{fig:1} (a), the system is not able to recognize the input audio: It responds as though the audio is not present.  
Furthermore, even when the system does recognize the existence of audio, it lacks the capability to decipher the information within the audio accurately. This is evident in Figure \ref{fig:1} (b) where all the sounds of people talking are disregarded as background noise, resulting in incorrect responses.
%
%
\begin{figure}[t]
  \centering
  \includegraphics[width=\linewidth]{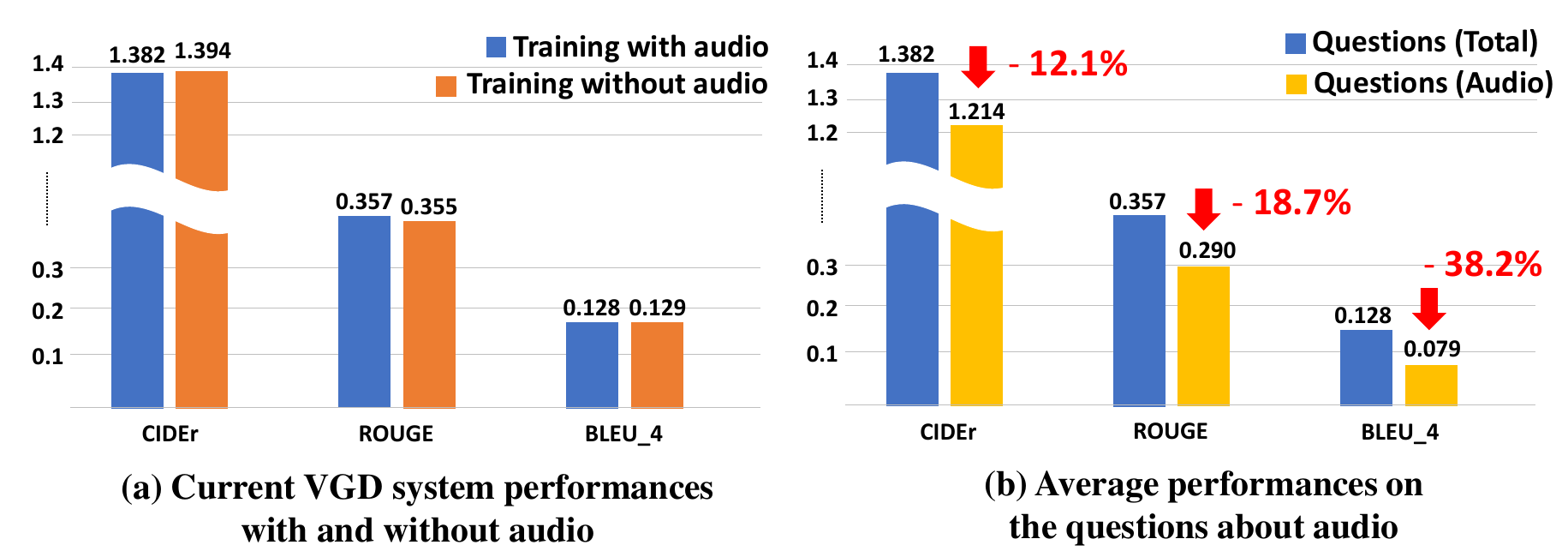}
  \caption{Current VGD systems' performances on AVSD dataset (validation): (a) Response performances according to training with and without audio, (b) Average performance drops on the questions about audio.}
  \label{fig:2}
  \vskip -0.2in
\end{figure}

%
Our experimental evidence in Figure \ref{fig:2} also shows that the current VGD systems cannot incorporate audio information when responding.
Figure \ref{fig:2} (a) shows the performance of response (\ie CIDEr \cite{vedantam2015cider}, ROUGE-L \cite{lin2004rouge}, BLEU \cite{papineni2002bleu}) of current VGD systems according to training with and without the audio: There are very little differences in performances between the two cases.
Some metrics even show a slightly higher performance when trained without audio.
%
%
Furthermore, as shown in Figure \ref{fig:2} (b), when investigating the responses to audio-related questions\footnote{We select the questions that contain words related to audio such as `sound', `speech', and `noise'.}, their performances are noticeably lower compared to the overall response performances.
Therefore, existing VGD systems tend to ignore the audio and suffer from the deaf response, which leads to incorrect answers, especially to questions pertinent to audio.

To overcome the deaf response problem, we propose Hearing Enhanced Audio Response (HEAR) framework, which allows the VGD systems to sensibly listen to audio according to the meaning of the question and perform enhanced listening to the audio.
%
%
Thus, HEAR incorporates (1) Sensible Audio Listening (SAL) that selectively attends to audio according to a sensible decision of whether to focus on audio or not and (2) Reconstructive Listening Enhancement (RLE) that improves the audibility via establishing a reconstruction upper bound to connect audio with its surrounding information.
For the sensible decision in SAL, we introduce two technical contributions: (1) Keyword-based Audio Sensing and (2) Semantic Neural Estimator.
HEAR is applied on current runner models \cite{Hori_2019_ICASSP,yoon-etal-2022-information,Li_2021_tip} in a model-agnostic manner, where the effectiveness is validated on VGD dataset (\ie AVSD@DSTC7, AVSD@DSTC8) with steady performance gains on natural language generation metrics.
%
%
%
%
%
%
%
%

\section{Related works}
\subsection{Video-grounded Dialogues}
Visual Question Answering (VQA) \cite{Antol_2015_ICCV,wang2022image} has been one of the proxy tasks to evaluate the multi-modal understanding of vision-language systems.
In light of recent advancements in generative models in natural language processing \cite{devlin2018bert,radford2018improving}, VQA has evolved into a more general format of answering as video-grounded dialogue (VGD) \cite{alamri2019audio}, where VGD aims to generate open-ended answer sentences from a question by referring to several input modalities (\ie video, audio, and dialogue history).
Many multi-modal interactions have been proposed, where various attention mechanisms \cite{Sanabria_2019_DSTC7,Le_2019_ACL} have been devised to perform cross-modal interactions.
To boost performances, transformer-based VGD systems \cite{Li_2021_tip} are utilized on top of large-scale pre-trained language models \cite{radford2019language,raffel2020exploring}.
%
%
Another immense challenge is keeping track of extended dialogue context, and video, where memory networks \cite{Lin_2019_DSTC7,Xie_2020_DSTC8} and multi-step attention \cite{Chu_2020_DSTC8} were introduced to efficiently store the video and long episodic dialogue.
Graph representations \cite{kim2021structured,pham2022video,le2021learning} were also popular solutions for holding semantic commonalities between the dialogue and video.
There have been novel structures \cite{Hori_2019_Interspeech,le2022vgnmn} to enhance the multi-modal representation and frameworks \cite{Le_2020_DSTC8,Lee_2020_DSTC8,yoon-etal-2022-information} to improve the quality of responses in terms of bias or word selection.
%
%
%
%
%
As such, many advances have been made in the multi-modal understanding of VGD systems, but mainly between video and language.
Thus, the VGD system's understanding of audio is still far from satisfactory.
To this end, we first contribute to improving the `listening' ability of VGD system.
%
\begin{figure*}[t]
	\centering
	\includegraphics[width=\linewidth]{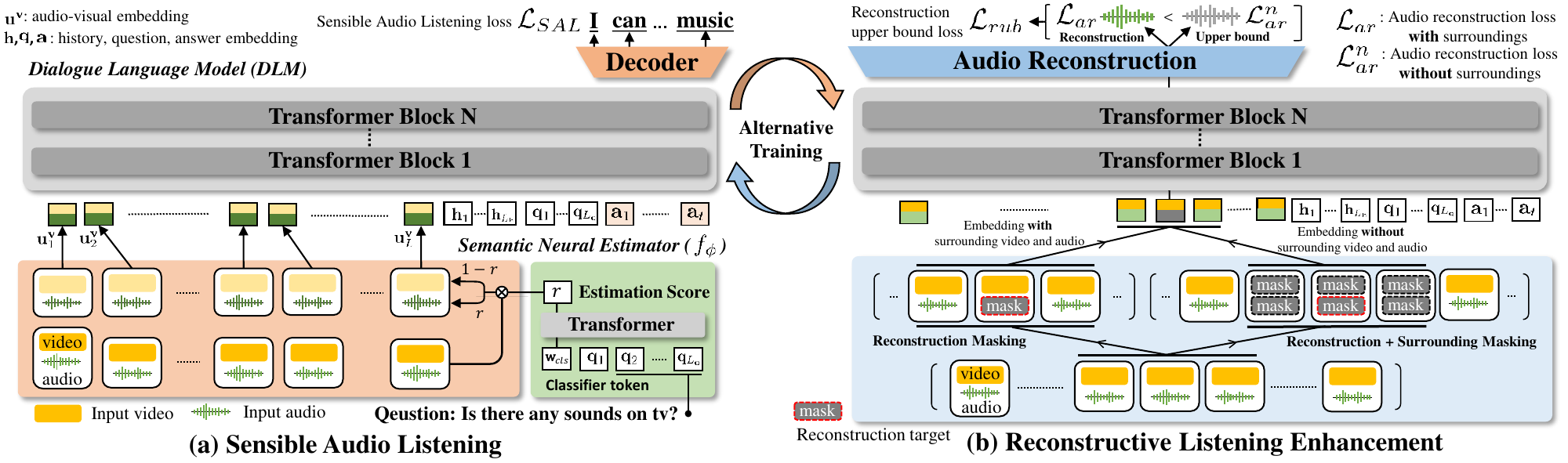}
	\caption{Illustration of Hearing Enhanced Audio Response Framework (HEAR) for video-grounded dialogue. HEAR performs sensible listening via (a) Sensible Audio Listening that selectively attends to audio corresponding to a given question and improves audibility via (b) Reconstructive Listening Enhancement that enhances audio representations by establishing a reconstruction upper bound to connect audio with its surrounding information.}
	\label{fig:model}
\end{figure*}
\section{Task Definition}
Video-grounded Dialogue \cite{alamri2019audio} (VGD) task aims to generate open-ended answer sentences to a question regarding multimodal inputs composed of video, audio, and dialogue history.
To build a formal task definition of VGD, a system takes tuples ($v,u,h,q^{r}$) as input and decodes answer sentence $a^{r}$, where $v$ is video, $u$ is audio, $h$ is dialogue history, and $q^{r}$ is a question asked at current round $r \in \{1,\cdots, R\}$.
The dialogue history $h = \{c,(q^{1},a^{1}),\cdots,(q^{r-1},a^{r-1})\}$ is composed of caption $c$ to summarize the video and a set of question-answer pairs in previous rounds. 
%
%
For training VGD system, next-word prediction is performed, where the system predicts $t$-th answer word token $a^{r}_{t}$ from the inputs of tuples ($v,u,h,q^{r}$) and partial answer word tokens $a^{r}_{<t}$ before $t$-th.

\section{Hearing Enhanced Audio Response}
Figure \ref{fig:model} illustrates Hearing Enhanced Audio Response (HEAR) framework designed to enhance Dialogue Language Model (DLM)\footnote{Here, we refer to the general `VGD system' as DLM.} in terms of two functionalities on audio: (1) Sensibility that selectively attends on audio according to the meaning of the question and (2) Audibility that performs enhanced listening to input audio.
For sensibility, we propose Sensible Audio Listening (SAL) in Figure \ref{fig:model} (a) which trains the DLM to respond to a question by selectively weighting to audio corresponding to the audio-relatedness of the question.
For audibility, we devise Reconstructive Listening Enhancement (RLE) in Figure \ref{fig:model} (b) which enhances audio representations by establishing a reconstruction upper bound to connect audio with its surrounding information.
We alternately train DLM with SAL and RLE to fully utilize video and audio modalities.
%
%
\subsection{Input representations}
We formally define input feature representations of $v$, $h$, $q^{r}$, and $a^{r}$ by embedding them into $d$-dimensional space. 
For the video embedding, we use I3D model \cite{carreira2017quo} pre-trained on the Kinetics dataset \cite{kay2017kinetics} to get 4096-dimensional video features $\mathbf{v} \in \mathbb{R}^{L \times 4096}$ composed of rgb and optical-flow features, where $L$ is the number of video frames.
%
%
%
%
For the audio embedding, we use VGGish model \cite{hershey2017cnn} pre-trained on the AudioSet dataset \cite{gemmeke2017audio} to get 128-dimensional audio features $\mathbf{u} \in \mathbb{R}^{L \times 128}$, where the $L$ is the number of audio features\footnote{Sample rate is modulated to be aligned with video frames.}.
The aforementioned video and audio features are concatenated along the feature dimension axis and embedded into $d$-dimensional space as audio-visual features $\mathbf{u}^{\mathbf{v}}$ as given below:
\begin{eqnarray}
\mathbf{u}^{\mathbf{v}} = [\mathbf{u}||\mathbf{v}]W \in \mathbb{R}^{L \times d},
\label{eq:1}
\end{eqnarray}
where $W \in \mathbb{R}^{(128+4096) \times d}$ is $d$-dimensional embbeder and $[\cdot||\cdot]$ denotes concatenation.

For the text features, we tokenize all the text inputs (\ie $h, q^{r}, a^{r}$) into a series of WordPieces \cite{wu2016google} using the T5 Transformer \cite{raffel2020exploring}, such that word token representations are obtained on top of relative positional embeddings and a layer normalization \cite{Ba_2016_arxiv}.
Thus, the formal definitions of text features are as follows: history $\mathbf{h} \in \mathbb{R}^{L_{\mathbf{h}} \times d}$, question $\mathbf{q} \in \mathbb{R}^{L_{\mathbf{q}} \times d}$, and answer $\mathbf{a} \in \mathbb{R}^{L_{\mathbf{a}} \times d}$, where $L_{\mathbf{h}}, L_{\mathbf{q}}$, and $L_{\mathbf{a}}$ are the numbers of tokens of each text\footnote{We delete superscripts $r$ in the notations for the simplicity.}.
\subsection{Dialogue Language Model}
Our proposed HEAR framework is performed in a model-agnostic manner, such that we first define Dialogue Language Model (DLM) as a general VGD system.
For input audio, video, and texts (\ie history and question), DLM is trained to generate next-word tokens for answer sentences $a^{r} = \{ a^{r}_{1},\cdots,a^{r}_{m}\}$ under cross-entropy loss as:
\begin{eqnarray}
\mathcal{L}_{DLM}(\theta) = -\textrm{log}\prod_{t=1}^{m} P_{\theta}(a^{r}_{t}|\mathbf{u}^{\mathbf{v}},\mathbf{h},\mathbf{q},\mathbf{a}_{<t}),
\end{eqnarray}
where $\theta$ is learnable weights and $m$ is the number of word tokens in the answer sentence. In the following, our proposed SAL and RLE improve the DLM's sensibility and audibility for correctly responding to questions about audio.
\subsection{Sensible Audio Listening}
\label{sec:4.3}
Sensible Audio Listening (SAL) is devised to determine whether the VGD system should listen to the audio to answer a given question. 
If the SAL determines that listening is required, audio is processed to be more weighted in the response.
To ensure sensible decision-making within SAL, we introduce two technical decision rules: (1) Keyword-based Audio Sensing and (2) Semantic Neural Estimator.
%
%
\paragraph{Keyword-based Audio Sensing.}
In many cases, unlike general questions, audio-related questions contain specific keywords (\eg `sound', `speech', `listen') that indicate that the question concerns audio.
Therefore,  we conduct an empirical investigation of these keywords.
If any of these keywords are present in a question, SAL identifies it as an audio-related question.
In such cases, we mask the video features in the inputs, directing more attention toward the audio component as given below:
%
%
%
%
\begin{eqnarray}
  \mathbf{u}^{\mathbf{v}}(q) = \begin{cases}
     [\mathbf{u}||\mathbf{v}]W & \text{$\forall$ $w_{q}$} \notin \text{W}_{key}\\
    [\mathbf{u}||\mathbf{m}_{\mathbf{v}}]W  & \text{otherwise},
  \end{cases}
  \label{eq:3}
\end{eqnarray}
where $w_{q}$ is all word tokens in the question $q$ and $\text{W}_{key}$ = $\{\text{`sound', `speech', `listen',}\dots\}$ is keyword set\footnote{Appendix provides all the keywords that we used.} to investigate audio-related questions. $\mathbf{m}_{\mathbf{v}} \in \mathbb{R}^{L \times 4096}$ is zero padding on video.
Here, we do not perform any masking when the question is not an audio-related question (\ie $\forall w_{q} \notin \text{W}_{key}$), because other questions excluding audio-related questions do not show high reliance on a specific modality.
Thus, $\mathbf{u}^{\mathbf{v}}(q) \in \mathbb{R}^{L \times d}$ is sensible audio-visual features that make DLM selectively focus on audio for responding to the audio-related questions. 
%
%
\begin{figure}[h!]
  \centering
  \includegraphics[width=\linewidth]{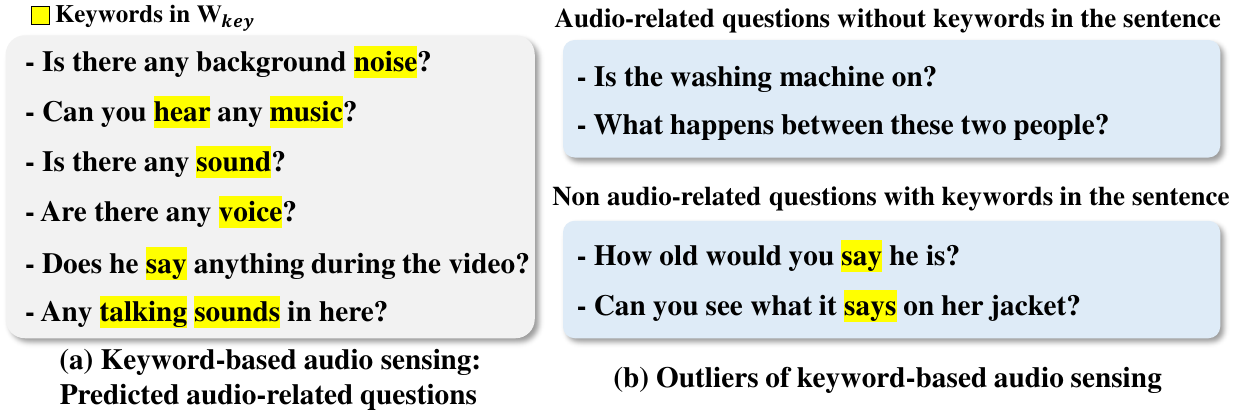}
  \caption{Examples of questions: (a) Predicted audio-related questions by keyword-based audio sensing and (b) Outliers of keyword-based audio sensing.}
  \label{fig:4}
\end{figure}
\paragraph{Semantic Neural Estimator.}
Figure \ref{fig:4} (a) illustrates audio-related questions identified by keyword-based audio sensing.
While the keyword-based approach effectively identifies the questions that directly require information about audio, we were still able to identify outliers that were not accurately identified in Figure \ref{fig:4} (b).
Although they do not contain keywords, the meanings of the questions are related to audio.
Therefore, we further devise a semantic neural estimator $f_{\phi}$ that identifies the audio-related question based on the meaning of sentences.
The semantic neural estimator as a BERT-based classifier takes $\{w_{\textrm{cls}},w_{q}\}$ to form an input instance, where $w_{\textrm{cls}}$ is a token for classification and the prediction target of $w_{\textrm{cls}}$ is $y \in \{y_{0},y_{1}\}$ denoting that $y_{1}=1$ is audio-related question and $y_{0}=0$ is the other question.
We first train the $f_{\phi}$ with training weights $\phi$ under L2 loss\footnote{Regularization is also used to mitigate training imbalance.} given as $\mathbb{E}_{D}(y - \hat{y})^{2}$, where $D$ is the dataset, $\hat{y}=f_{\phi}(w_{\textrm{cls}},w_{q})$ is prediction score of the sigmoid (\ie $0<\hat{y}<1$). 
%
For labeling the $y$, we first include the predictions by keyword-based audio sensing as a noisy label. 
We further include $y_{0}$ with (1) word-wise random shuffled versions of audio-related questions and (2) random swapping versions of non-audio questions using keyword $\text{W}_{key}$. 
This prevents $f_{\phi}$ from simply predicting based on the keywords and makes more focus on the meaning of the audio-related questions in $y_{1}$.
%
%
After training $f_{\phi}$, we calibrate the audio and video features according to the estimation of $f_{\phi}$ as:
\begin{eqnarray}
\mathbf{u}^{\mathbf{v}}(q) = [ r \times  \mathbf{u}|| (1-r) \times \mathbf{v}]W,
\end{eqnarray}
%
%
where $r=f_{\phi}(w_{\textrm{cls}},w_{q}) \in \mathbb{R}$ is estimation score about audio-related question and $\mathbf{u}^{\mathbf{v}}(q) \in \mathbb{R}^{L \times d}$ is our final sensible audio-visual features for SAL.
%
%
%
%
Based on the sensible audio-visual feature $\mathbf{u}^{\mathbf{v}}(q)$ from SAL, we train Dialogue Language Model (DLM) with cross-entropy loss as given below:
\begin{eqnarray}
\mathcal{L}_{SAL}(\theta) = -\textrm{log}\prod_{t=1}^{m} P_{\theta}(a^{r}_{t}|\mathbf{u}^{\mathbf{v}}(q),\mathbf{h},\mathbf{q},\mathbf{a}_{<t}).
\end{eqnarray}
\begin{figure}[t]
	\centering
	\includegraphics[width=\linewidth]{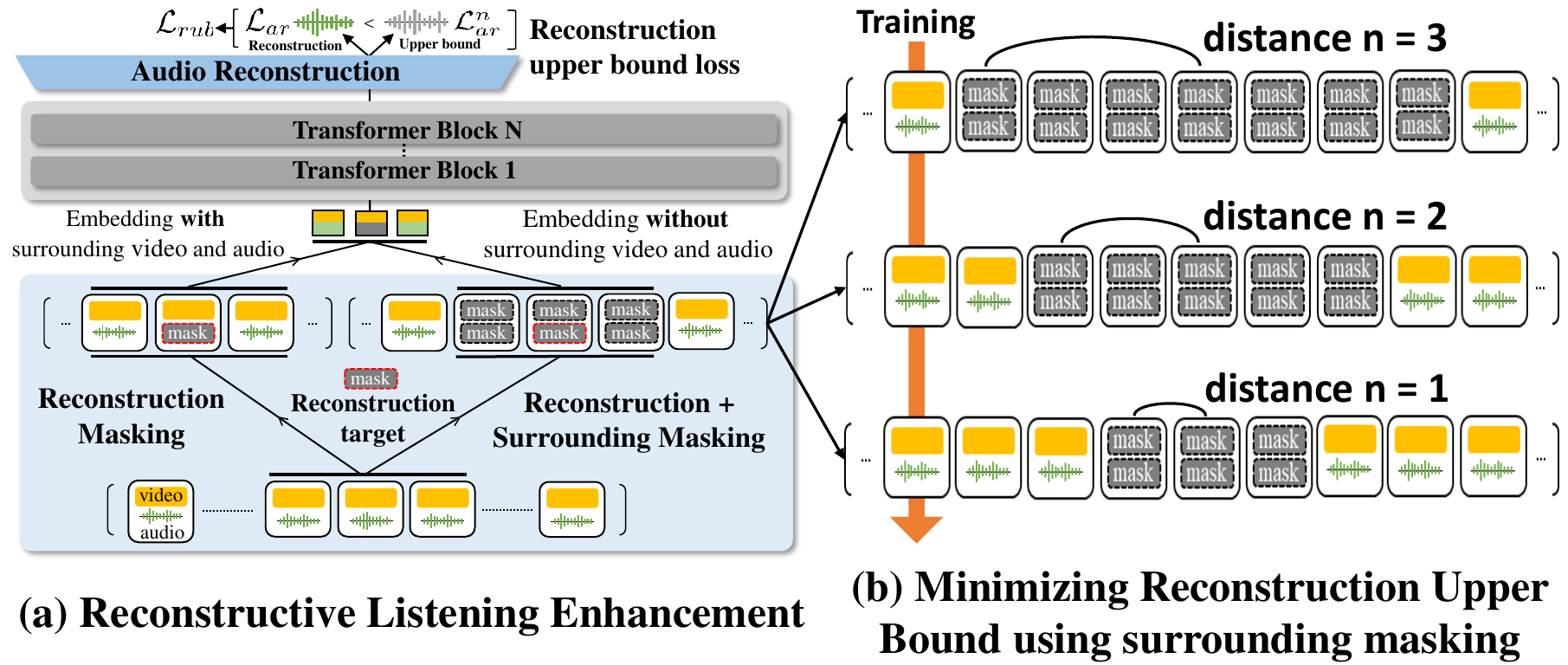}
	\caption{Illustrations of surrounding masking. The distance $n$ decides the extent of surrounding masking.}
	\label{fig:sc}
\end{figure}
\subsection{Reconstructive Listening Enhancement}
\label{sec:4.4}
For a better understanding of audio, it is crucial to improve audio representations based on a common understanding of a scene in a video.
Therefore, our proposed Reconstructive Listening Enhancement (RLE) performs masked audio reconstruction by referring to surrounding information (\ie video and audio adjacent to masked audio), which enhances the common embedding with its surroundings.
Especially to perform effective enhancement in regions closer to the masked target, we propose Reconstruction Upper Bound in the following.
%
%
%
\paragraph{Audio Reconstruction.}
%
%
%
Audio reconstruction aims to predict masked audio based on observations of their surrounding audio and other modalities (\ie video and text).
We randomly select input audio with a probability of $p$ (\eg $p$=10\%) as $\mathbf{u}_{m}$, where $\mathbf{u}_{m} \in \mathbb{R}^{M \times 128}$ is the target audio features to be masked, $m$ is the set of indices for masking, and $M$ is the number of indices.
We also define $\mathbf{u}_{\backslash m} \in \mathbb{R}^{L \times 128}$ as surrounding audio features including masked features of zero vectors in the indices $m$. 
%
%
We introduce audio reconstruction loss $\mathcal{L}_{ar}$ for DLM to reconstruct target audio features $\mathbf{u}_{m}$ from the inputs of $\{\mathbf{u}_{\backslash m}, \mathbf{v}, \mathbf{h},\mathbf{q}\}$ as:
\begin{equation}
\begin{aligned}
\mathcal{L}_{ar}(\theta) = h_{\theta}(\mathbf{u}_{m}|\mathbf{u}_{\backslash m}, \mathbf{v}, \mathbf{h},\mathbf{q}).
\end{aligned}
\end{equation}
$h_{\theta}(\mathbf{u}_{m}|\mathbf{u}_{\backslash m}, \mathbf{v}, \mathbf{h},\mathbf{q}) =\sum_{i=1}^{M}||\mathbf{u}_{m}^{(i)}- \mathbf{\tilde{u}}_{m}^{(i)} ||_{2}^{2}$ is l2 regression, where $\mathbf{\tilde{u}}_{m} \in \mathbb{R}^{M \times 128}$ is reconstructed audio on the masked indices $m$ from the output of multi-layer perceptron on top of DLM encoder.
%
%
%
%
%
\paragraph{Reconstruction Upper Bound.}
The audio reconstruction should be based on an understanding of surrounding semantics rather than simply memorizing audio.
To facilitate this, we propose Reconstruction Upper Bound (RUB) which establishes an inequality condition to ensure enhanced reconstruction when the surrounding information is provided compared to when it is not given as given below:
%
\begin{eqnarray}
\mathcal{L}_{ar}(\theta) < h_{\theta}(\mathbf{u}_{m}|\mathbf{u}_{\backslash m}^{n}, \mathbf{v}^{n},\mathbf{h},\mathbf{q}),
\end{eqnarray}
where the audio reconstruction loss based on surroundings $\mathcal{L}_{ar}(\theta) = h_{\theta}(\mathbf{u}_{m}|\mathbf{u}_{\backslash m}, \mathbf{v},\mathbf{h},\mathbf{q})$ should be lower than the reconstruction loss without surroundings $\mathcal{L}_{ar}^{n}(\theta) = h_{\theta}(\mathbf{u}_{m}|\mathbf{u}_{\backslash m}^{n}, \mathbf{v}^{n},\mathbf{h},\mathbf{q})$ (\ie upper bound).
As shown in Figure \ref{fig:sc} (a), $\mathbf{u}_{\backslash m}^{n}, \mathbf{v}^{n}$ are audio and video features that further mask out surroundings up to feature distance $n$ (\eg $n$ = 3) from the target audio features.
To train DLM to conform this inequality condition, we calculate a reconstruction upper bound loss $\mathcal{L}_{rub}$ in a format of ranking loss with margin $\delta$ as given below: 
\begin{eqnarray}
  \mathcal{L}_{rub}(\theta,n) = \textrm{max}(\mathcal{L}_{ar}(\theta) - \mathcal{L}_{ar}^{n}(\theta) + \delta, 0).
\end{eqnarray}
\begin{figure}[t]
	\centering
	\includegraphics[width=\linewidth]{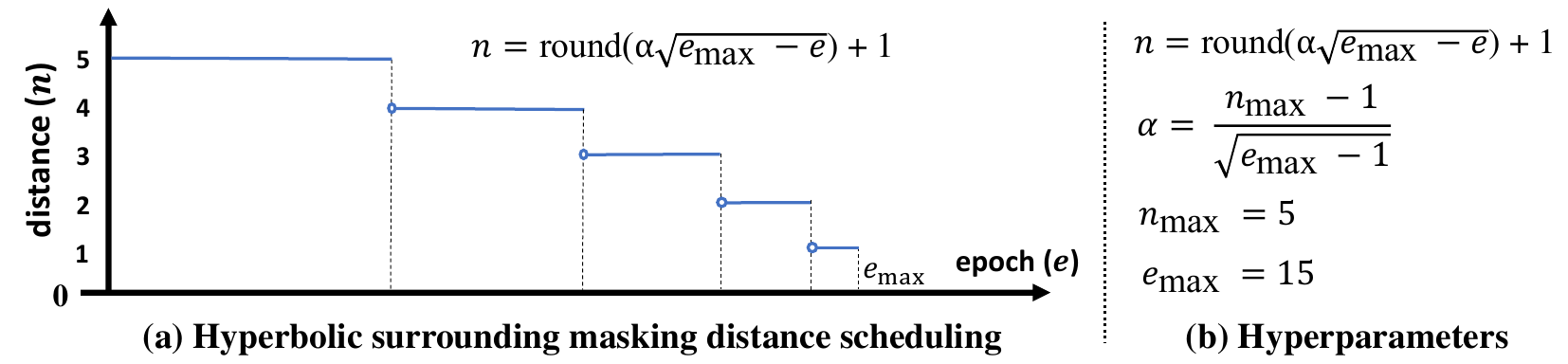}
	\caption{Hyperbolic function for distance scheduling.}
	\label{fig:hp}
\end{figure}

After that, we iteratively minimize the upper bound $\mathcal{L}_{ar}^{n}$ in a way of narrowing the surrounding masking with distance $n$\footnote{As the surrounding masking is reduced, the DLM can perform the reconstruction better, thus $\mathcal{L}_{ar}^{n}$ decreases.} as shown in Figure \ref{fig:sc} (b). 
Here, we construct masking distance scheduling $n = g(e)$ according to taring epoch $e$ to select progressively lower $n$ from higher $n$\footnote{Empirically found that low $n$ in early training (\eg S(e) = 1) ruins training stability, as $\mathcal{L}_{ar}$, $\mathcal{L}_{ar}^{n}$ were almost similar.}.
%
%
%
Therefore, the final objective of RLE is formulated as below:
\begin{eqnarray}
 \underset{\theta,g\sim G}{\textrm{min}} \mathcal{L}_{ar}(\theta) +  \mathcal{L}_{rub}(\theta,g(e)),
\end{eqnarray}
where $G:\mathbb{R}^{+} \mapsto \mathbb{R}^{+}$ is a set of scheduling functions and we select hyperbolic function\footnote{See more scheduling functions in Appendix B.2.} as $g(e) = \textrm{round}(\alpha \sqrt{e_{\textrm{max}} - e}) + 1$ as depicted in Figure \ref{fig:hp} to give stable decreasing of $n$ with $\alpha=\frac{n_{\textrm{max}}-1}{\sqrt{e_{\textrm{max}}-1}}$ and $n_{\textrm{max}}$=5, $e_{\textrm{max}}$=15 are maximum distance and epoch.
Therefore, the final RLE loss is the summation of two losses: $\mathcal{L}_{RLE} = \mathcal{L}_{ar} +  \mathcal{L}_{rub}$.
%
%
%
\begin{table*}[t]
	\centering
	\begin{tabular}{lc c c c c c c}
		\Xhline{3\arrayrulewidth}
		\multicolumn{8}{c}{AVSD@DSTC7}\\ \Xhline{3\arrayrulewidth}
		Methods                              & B1     & B2     &B3     & B4     & M   & R   & C     \\ 
		\Xhline{2\arrayrulewidth}
		EE-DMN \cite{Lin_2019_DSTC7}         & 0.641     & 0.493     & 0.388     & 0.310     & 0.241    & 0.527     & 0.912     \\
		JMAN \cite{Chu_2020_DSTC8}           & 0.667     & 0.521     & 0.413     & 0.334     & 0.239    & 0.533     & 0.941     \\
		CMU \cite{Sanabria_2019_DSTC7}       & 0.718     & 0.584     & 0.478     & 0.394     & 0.267    & 0.563     & 1.094     \\
        COST \cite{pham2022video} &	0.723 & 0.589 & 0.483 & 0.400 &0.266 &0.561 &1.085 \\
		MSTN \cite{Lee_2020_DSTC8}           & -         & -         & -         & 0.377     & 0.275    & 0.566     & 1.115     \\
		JSTL \cite{Hori_2019_Interspeech}    & 0.727     & 0.593     & 0.488     & 0.405     & 0.273    & 0.566     & 1.118     \\
		MTN$^{\star}$ \cite{Le_2019_ACL}               & 0.731     & 0.597     & 0.490     & 0.406     & 0.271    & 0.564     & 1.127     \\
		MTN-P \cite{Le_2020_DSTC8}           & 0.750& 0.619     & 0.514     & 0.427     & 0.280    & 0.580& 1.189     \\
		VGNMN \cite{le2022vgnmn}                     & -     & -     & -     & 0.429     & 0.278    & 0.578     & 1.188     \\
		SCGA \cite{kim2021structured}                                 & 0.745     & 0.622& 0.517& 0.430&0.285& 0.578     & 1.201\\	
		RLM \cite{Li_2021_tip} &	0.765     & 0.643 & 0.543 & 0.459 &0.294 &0.606 &1.308\\
          PDC \cite{le2021learning} & 0.770 & 0.653 &0.539 & 0.449 & 0.292 & 0.606 & 1.295 \\
		THAM \cite{yoon-etal-2022-information}                 & 0.778 &0.654 &0.549 &0.468 &0.308 &0.619 &1.335\\	\hline
  \textbf{HEAR (SAL)}                                 & 0.784 &0.658 &0.551 &0.471 &0.309 &0.619 &1.347\\
  \textbf{HEAR (SAL + RLE)}                                 & \bf{0.791} &\bf{0.662} &\bf{0.558} &\bf{0.472} &\bf{0.312} &\bf{0.622} &\bf{1.376}\\	
		\Xhline{3\arrayrulewidth} 
		\multicolumn{8}{c}{AVSD@DSTC8}\\ \Xhline{3\arrayrulewidth}
		MDMN \cite{Xie_2020_DSTC8}           & -       & -       & -       & 0.296   & 0.214  & 0.496   & 0.761 \\
		JMAN \cite{Chu_2020_DSTC8}           & 0.645   & 0.504   & 0.402   & 0.324   & 0.232  & 0.521   & 0.875 \\
		STSGR \cite{geng2021dynamic}         & -       & -       & -       & 0.357   & 0.267  & 0.553   & 1.004 \\
		MSTN \cite{Lee_2020_DSTC8}           & -       & -       & -       & 0.385   & 0.270  & 0.564   & 1.073 \\
        COST \cite{pham2022video} &	0.695     & 0.559 & 0.465 & 0.382 &0.278 &0.574 &1.051\\
		MTN-P \cite{Le_2020_DSTC8}           & 0.701   & 0.587   & 0.494   & 0.419   & 0.263  & 0.564   & 1.097 \\
		SCGA \cite{kim2021structured}                                 & 0.711   & 0.593   & 0.497 & 0.416   & 0.276  & 0.566   & 1.123 \\ 
		RLM \cite{Li_2021_tip}                                 & 0.746     & 0.626 & 0.528 & 0.445 &0.286 &0.598     & 1.240\\ 
        PDC \cite{le2021learning} & 0.749 & 0.629 &0.528 & 0.439 & 0.285 & 0.592 & 1.201 \\
		THAM \cite{yoon-etal-2022-information}                                &0.764     &0.641 &0.538 &0.455 &0.301 &0.610 &1.304\\ \hline	    
  \textbf{HEAR (SAL)}                                 &0.767     &0.646 &0.541 &0.459 &0.302 &0.612 &1.324\\	
  \textbf{HEAR (SAL + RLE)}                                 &\textbf{0.777}     &\textbf{0.656} &\textbf{0.553} &\textbf{0.465} &\textbf{0.307} &\textbf{0.618} &\textbf{1.359}\\	
		\Xhline{3\arrayrulewidth}
	\end{tabular}
    \caption{Experimental results on AVSD@DSTC7 (test) and AVSD@DSTC8 (test) dataset. (B: BLEU, M: METEOR, R: ROUGE-L, C: CIDEr, $\star$: reported in \cite{kim2021structured}). Human Evaluation is also presented in Appendix C.}
	\label{tab:1}
\end{table*}
\subsection{Optimization and Inference}
As shown in Figure \ref{fig:model}, we train DLM by alternately optimizing $\mathcal{L}_{SAL}$ and $\mathcal{L}_{RLE}$ to allow audio and video to be freely utilized for their purposes.
Therefore, $\mathcal{L}_{SAL}$ is optimized when the training iteration number is odd, and $\mathcal{L}_{RLE}$ does in the even number:
%
%
\begin{eqnarray}
  \mathcal{L}_{HEAR} = \begin{cases}
     \mathcal{L}_{SAL} & \text{iteration} = odd\\
    \mathcal{L}_{RLE}  & \text{iteration} = even,
  \end{cases}
\end{eqnarray}
Here, RLE is only used for training. In an inference, The DLM generates an answer with SAL. 
%
%
\section{Experiments}
\subsection{Datasets}
\paragraph{AVSD@DSTC7 and AVSD@DSTC8.} (Audio Visual Scene-Aware Dialog) \cite{alamri2019audio,hori2020audio} is a popular benchmark for VGD task, where all videos include their summary captions and dialogue composed of 10 pairs of question and answer.
The videos are collected from Charades \cite{sigurdsson2016hollywood} dataset, which contains natural human activities.
AVSD is released at Dialogue System Technology Challenge (DSTC) 7 and 8, where AVSD@DSTC7 contains $7,659$, $1,787$, and $1,710$ dialogues for training, validation and test, but AVSD@DSTC8 only provides $1,710$ dialogues for the test.
For the test set, $6$ reference answers are available for accurate validation.
\begin{table*}[t]
	\centering
	\begin{tabular}{lc c c c c c c}
		\Xhline{3\arrayrulewidth}
		\multicolumn{8}{c}{AVSD@DSTC7}\\ \Xhline{3\arrayrulewidth}
		Methods                              & B1     & B2     &B3     & B4     & METEOR   & ROUGE-L   & CIDEr 
   \\ 
		\Xhline{2\arrayrulewidth}
		$\textrm{MTN}^{\dagger}$ \cite{Le_2019_ACL}         & 0.357 & 0.241 & 0.173 & 0.128 & 0.162  & 0.355 & 1.249      \\
		\bf{$\textrm{MTN}^{\dagger}$ + HEAR}            & 0.382 & 0.264 & 0.193 & 0.144 & 0.179  & 0.374 & 1.314    \\ \hline
		RLM \cite{Li_2021_tip}                              & 0.765 & 0.643 & 0.543 & 0.459 & 0.294  & 0.606 & 1.308 \\
		\bf{RLM + HEAR}                                 & 0.787 & 0.660 & 0.556 & 0.467 & 0.308 & 0.619 & 1.358  \\\hline
		T5RLM \cite{yoon-etal-2022-information}                                   & 0.767 &0.644  &0.542 &0.461 &0.296 &0.608 &1.311  \\	
		\bf{T5RLM + HEAR}                               & \bf{0.791} &\bf{0.662} &\bf{0.558} &\bf{0.472} &\bf{0.312} &\bf{0.622} &\bf{1.376} \\ \Xhline{3\arrayrulewidth}
  		\multicolumn{8}{c}{AVSD@DSTC8}\\ \Xhline{3\arrayrulewidth}
		$\textrm{MTN}^{\star}$ \cite{Le_2019_ACL}          & 0.689   & 0.571   & 0.470   & 0.404   & 0.251  & 0.551   & 1.049 \\
		\bf{$\textrm{MTN}^{\star}$ + HEAR}          & 0.714   & 0.596   & 0.496   & 0.421   & 0.271  & 0.569   & 1.121 \\ \hline
		RLM \cite{Li_2021_tip}                                 & 0.746     & 0.626 & 0.528 & 0.445 &0.286 &0.598     & 1.240 \\
		\bf{RLM + HEAR}                                 & 0.772     & 0.651 & \bf{0.554} & 0.462 &0.303 &0.617     & 1.323\\ \hline
		T5RLM \cite{yoon-etal-2022-information}                                  & 0.749 &0.631 &0.529 &0.445 &0.290 &0.600 &1.263\\
		\bf{T5RLM+ HEAR}                                 &\bf{0.777}     &\bf{0.656} &0.553 &\bf{0.465} &\bf{0.307} &\bf{0.618} &\bf{1.359} \\	
		\Xhline{3\arrayrulewidth}
	\end{tabular}
 \caption{Experimental results on AVSD@DSTC7 (test) and AVSD@DSTC8 (test) for applying HEAR on VGD runner models (B1: BLEU1, $\star$: reconstruction-based results, $\dagger$: single reference results).}
	\label{tab:2}
\end{table*}
%
\begin{table}
  \centering
  \small
  \begin{tabular}{ccccccc}
  \toprule
    \multicolumn{2}{c}{SAL} &\multicolumn{2}{c}{RLE} & \multicolumn{1}{c}{\multirow{2}{*}{BLEU1}} & \multicolumn{1}{c}{\multirow{2}{*}{ROUGE-L}} & \multicolumn{1}{c}{\multirow{2}{*}{CIDEr}} \\ \cmidrule(rl){1-2} \cmidrule(rl){3-4}
      k & s & $\mathcal{L}_{ar}$ & $\mathcal{L}_{rub}$ & & \\ \midrule
      &  &  &    &0.302  &0.348 &1.367 \\ \cmidrule(rl){1-7}
    \checkmark  &  &   &     &0.310  &0.355 &1.427 \\
      & \checkmark &   &    &0.316  &0.366 &1.434\\ 
      & \checkmark & \checkmark  &    &0.329  &0.375 &1.467\\
      & \checkmark &             & \checkmark   &0.311  &0.357 &1.415\\ 
      & \checkmark & \checkmark  & \checkmark  &\bf{0.341}  &\bf{0.384} &\bf{1.492}\\ \bottomrule
  \end{tabular}
  \caption{Ablation studies of model variants on AVSD@DSTC7 (validation, single reference). k: Keyword-based Audio Sensing, s: Semantic Neural Estimator, $\mathcal{L}_{ar}$: Audio Reconstruction, $\mathcal{L}_{rub}$: Reconstruction Upper Bound}
  \label{tab:ablation_loss}
\end{table}
\subsection{Metrics}
We follow official metrics for AVSD benchmark (\ie BLEU, METEOR \cite{banerjee2005meteor}, ROUGE-L, CIDEr) provided by AVSD challenge organizers\footnote{\textrm{github.com/dialogtekgeek/DSTC8-AVSD\_official}}, where they compute words overlaps between predicted sentence and reference answer.
\subsection{Results on AVSD benchmark}
Table \ref{tab:1} summarizes the experimental results of HEAR on the AVSD@DSTC7 and AVSD@DSTC8.
HEAR shows state-of-the-art performances on all the metrics compared to previous works (\ie please refer to Related Work for their detailed captions.). 
Our baseline DLM is T5 Transformer \cite{raffel2020exploring}, which is the same baseline (\ie T5RLM) of THAM \cite{yoon-etal-2022-information}, but here, our proposed SAL shows more gains, and further improvements are also shown by applying RLE.
As our proposed HEAR is performed in a model-agnostic manner, we also validate other VGD models with HEAR in Table \ref{tab:2}.
We utilize the public codes and papers for MTN, RLM, and T5RLM, where steady gains in all metrics are shown for all the models.
%
%
%
\begin{figure}[t]
  \centering
  \includegraphics[width=\linewidth]{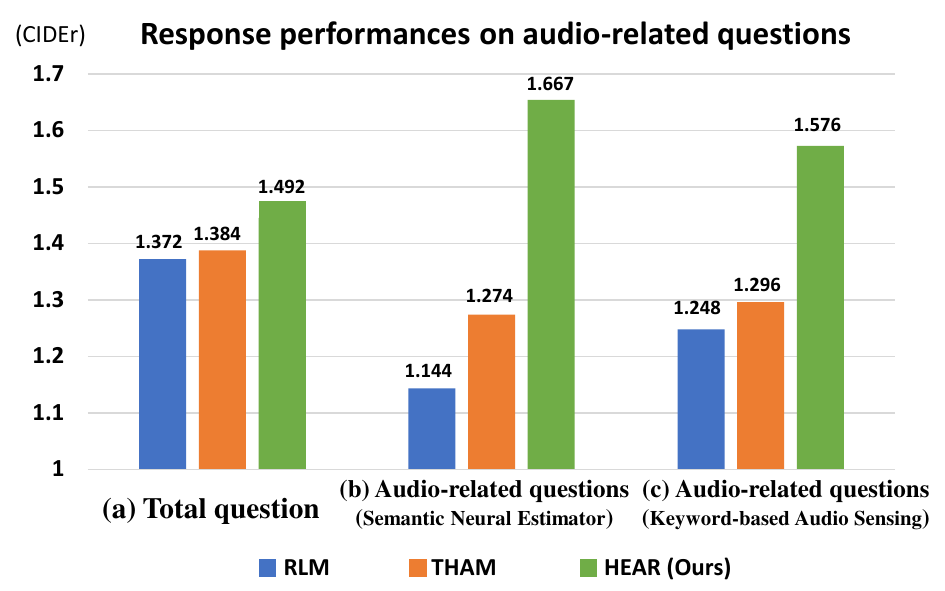}
  \caption{VGD systems' response performances on audio-related questions of AVSD@DSTC7 (validation): (a) Total questions, (b) Audio-related questions predicted by Semantic Neural Estimator (questions with estimation score $r>0.7$), (c) Audio-related questions predicted by Keyword-based Audio Sensing.}
  \label{fig:5}
\end{figure}
\subsection{Ablation Study}
\label{ablation}
Table \ref{tab:ablation_loss} summarizes the ablative results of the proposed modules in HEAR framework.
The first section of Table \ref{tab:ablation_loss} is about our base DLM performances.
In the second section, for the variants of SAL, smaller gains are obtained when only using keyword-based audio sensing.
We think that using it alone was not beneficial in answering some audio-related questions that require referencing the videos, as this method unconditionally screens out video features as long as the given questions are related to audio.
In the case of RLE, the audio reconstruction loss $\mathcal{L}_{ar}$ generally gives a positive effect on the system performances. 
However, the reconstruction upper bound loss $\mathcal{L}_{rub}$ becomes effective only when the $\mathcal{L}_{ar}$ and $\mathcal{L}_{rub}$ are used together.
%
This denotes that the ranking loss in the $\mathcal{L}_{rub}$ is founded on well-reconstructed audio features.
%

One may wonder if HEAR really helps answer audio-related questions.
Figure \ref{fig:5} shows the response performances in terms of VGD systems including HEAR according to questions.
HEAR performs new state-of-the-art performances on top of previous runner models, but it is also meaningful that the gains are mainly obtained from improving responses to audio-related questions.
From the results in Figure \ref{fig:5} (b), (c) and Table \ref{tab:2}, it is concluded that HEAR framework can be applied to any VGD system, where it exclusively contributes to improving systems' audibility and generating correct responses to given audio-related questions.
\begin{figure}[t]
  \centering
  \includegraphics[width=\linewidth]{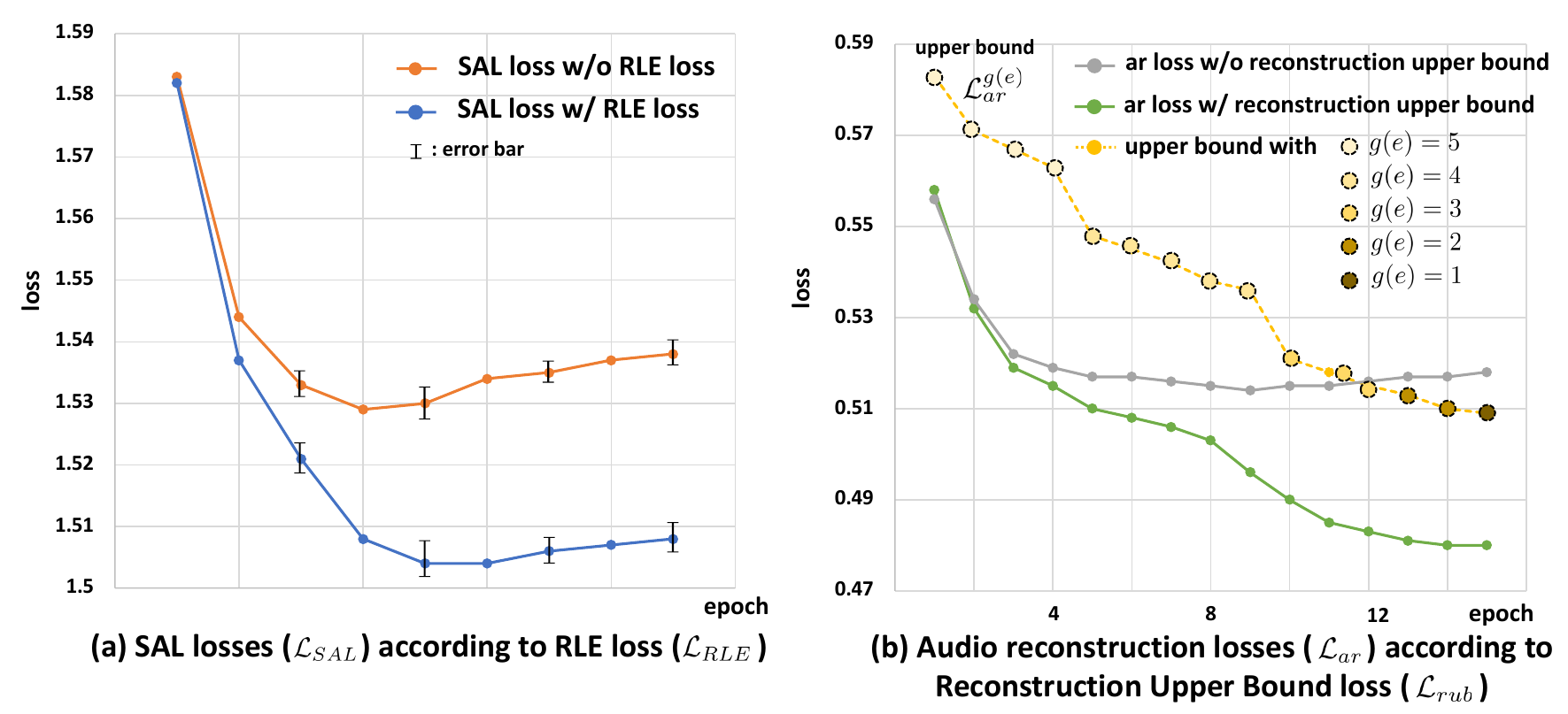}
  \caption{Ablative results on validation losses according to epochs: (a) SAL loss $\mathcal{L}_{SAL}$ with and without RLE loss $\mathcal{L}_{RLE}$, (b) Audio reconstruction loss $\mathcal{L}_{ar}$ with and without reconstruction upper bound loss $\mathcal{L}_{rub}$.}
  \label{fig:6}
\end{figure}

Furthermore, we found that our proposed RLE also contributes to the efficient learning of sensible audio listening (\ie optimizing $\mathcal{L}_{SAL}$).
Figure \ref{fig:6} (a) shows validation losses ($\mathcal{L}_{SAL}$) according to training epochs, where it can be seen that the $\mathcal{L}_{SAL}$ is further reduced when RLE loss ($\mathcal{L}_{RLE}$) is applied.
%
%
%
Figure \ref{fig:6} (b) shows the ablation studies of audio reconstruction loss $\mathcal{L}_{ar}$ with and without reconstruction upper bound $\mathcal{L}_{rub}$. 
The yellow curve shows the upper bound loss $\mathcal{L}_{ar}^{g(e)}$ for $\mathcal{L}_{rub}$, where the it decreases according to our masking distance scheduling $g(e)$\footnote{We use a hyperbolic curve for $g(e)$. Appendix also contains more diverse surrounding masking scheduling.}. 
The green curve denotes the $\mathcal{L}_{ar}$ with $\mathcal{L}_{rub}$ and it shows further optimizing compared to without $\mathcal{L}_{rub}$ as the $\mathcal{L}_{ar}^{g(e)}$ decreases. 
This denotes that neural networks can be further optimized according to the training epochs by calibrating their training objectives, which is also validated in other multi-modal systems \cite{yoon2023scanet,zheng2022weakly} in other ways.
%
%
%
%
\begin{figure}[t]
  \centering
  \includegraphics[width=\linewidth]{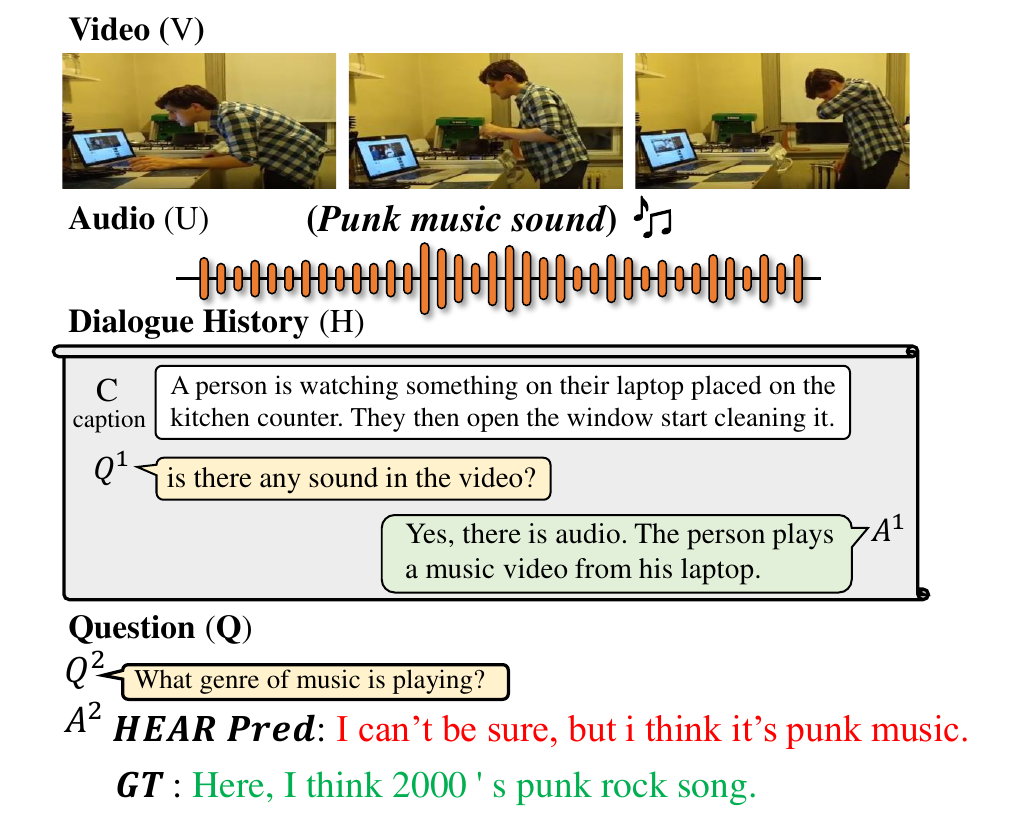}
  \caption{HEAR responses to an audio-related question.}
  \label{fig:7}
\end{figure}
\begin{table}
  \centering
  \small
  \begin{tabular}{lccc}
  \toprule
   Questions & K & S & H\\     \midrule
   (a) \textit{Can you hear any sounds?} & $T$ & $0.99$ & $T$\\ 
   (b) \textit{Do they speak to each other?} & $T$ & $0.99$ & $T$\\ 
   (c) \textit{Who is outside the door?} & $F$ & $0.77$ & $T$\\ 
   (d) \textit{Is the vacuum cleaner working?} & $F$ & $0.74$ & $T$\\ 
   (e) \textit{Can you tell where he goes?} & $T$ & $0.32$ & $F$\\ 
   (f) \textit{What color is his hair?} & $F$ & $0.01$ & $F$\\ \bottomrule
  \end{tabular}
  \caption{Predictions on questions about audio relatedness. K: Keyword-based Audio Sensing, S: Semantic Neural Estimator, H: human rating, $T$: true, $F$: false.}
  \label{tab:audio_que}
\end{table}
\subsection{Qualitative Results}
Figure \ref{fig:7} illustrates the HEAR's responses to audio-related question.
HEAR is given the question ``what genre of music is playing?'' as the audio-related question, and it generates the answer sentence ``I can't be sure, but I think it's punk music.''
Here, HEAR precisely predicted that the given audio was punk music despite the challenging work of discerning what the sound is in the video. 
The more interesting fact is that HEAR represents its opinion as ``I can't be sure'' about predicting music. 
When we listened to that audio in Figure \ref{fig:7} (a), it was really quiet and difficult to identify what the sound is.
We guess the HEAR also learned the knowledge about which sounds are difficult for humans to distinguish.
Table \ref{tab:audio_que} shows predictions on questions by Keyword-based Audio Sensing and Semantic Neural Estimator.
As the keyword-based approach could not understand the meaning of the question, it shows incorrectness in some questions (\ie (c,d,e)).
The neural estimator presents the score $0<r<1$ denoting whether the question is related to the audio, which provides proper distinctions between audio-related questions (\ie (a,b,c,d)) and the others (\ie (e,f)) based on the meaning of the question.
%
%
%
\section{Conclusion}
%
%
We propose Hearing Enhanced Audio Response (HEAR) framework for Video-grounded Dialogue systems.
HEAR proposes Sensible Audio Listening and Reconstructive Listening Enhancement, which solve the deaf response problem of VGD systems and validate model-agnostic effectiveness on top of current VGD systems.
%
%
\section*{Acknowledgements}
This work was supported by Institute for Information \& communications Technology Promotion(IITP) grant funded by the Korea government(MSIT) (No. 2021-0-01381, Development of Causal AI through Video Understanding and Reinforcement Learning, and Its Applications to Real Environments) and partly supported by a grant of the KAIST-KT joint research project through AI2XL Laboratory, Institute of convergence Technology, funded by KT [Project No. G01220646, Visual Dialogue System: Developing Visual and Language Capabilities for AI-Based Dialogue Systems].
%
%
\section*{Limitations}
Our research aims to enhance the audibility of Video-grounded Dialogue (VGD) systems, where we devise Hearing Enhanced Audio Response (HEAR) Framework.
As a limitation of our work, we are more focused on the methodology. We need to have a better understanding of the limitation of the proposed method to overcome any failure cases.
%
%
As shown in the failure case in Appendix, our proposed method has limitations, and we would need to consider other architecture and training methods to incorporate all the necessary information that speech holds.
To understand human speech, current audio features seem necessary to be trained more by large-scale audio speech recognition datasets \cite{panayotov2015librispeech,garofolo1993timit}.
Furthermore, although our proposed HEAR mitigates this problem in a methodological idea, we also think that the deaf problem can also be cured by expanding the audio feature dimension (\ie 128) up to a comparable scale with video (\ie 4096), such that they include more detailed information. 
To be specific, we are currently changing the current audio feature extractor (\ie VGGish) into wav2vec 2.0 \cite{baevski2020wav2vec}, which can provide a larger dimensional audio feature (\ie 768 dimension).
We will also make the audio features (\ie wav2vec 2.0 features) publicly available and perform a further study on this as our future work.
%
%
%
%
\section*{Ethics Statement}
Video-grounded Dialogue system is one of the conversational AI agents, which is designed to provide assistance to various subsections of our environments including security, entertainment, education, and visual impairments.
Our proposed Hearing Enhanced Audio Response framework contributes to improving responses to queries about audio by enhancing the audibility of VGD system.
Recently chatbot systems (\eg ChatGPT) has shown overwhelming performance, as such, we should also think about the potential negative societal impact of these systems.
In this respect, we came up with two negative impacts:  (1) unreliable vague information by conversational agents and (2) fairness issues in agents' responses.
Therefore, word sense disambiguation techniques \cite{yoon2022smsmix} and multi-modal debiasing solutions \cite{yoon2022selective,niu2021counterfactual} should also be applied to the dialogue systems.
%

\bibliography{anthology,custom}

\begin{thebibliography}{46}
\expandafter\ifx\csname natexlab\endcsname\relax\def\natexlab#1{#1}\fi

\bibitem[{Alamri et~al.(2019)Alamri, Cartillier, Das, Wang, Cherian, Essa, Batra, Marks, Hori, Anderson et~al.}]{alamri2019audio}
Huda Alamri, Vincent Cartillier, Abhishek Das, Jue Wang, Anoop Cherian, Irfan Essa, Dhruv Batra, Tim~K Marks, Chiori Hori, Peter Anderson, et~al. 2019.
\newblock Audio visual scene-aware dialog.
\newblock In \emph{Proceedings of the IEEE/CVF Conference on Computer Vision and Pattern Recognition}, pages 7558--7567.

\bibitem[{Antol et~al.(2015)Antol, Agrawal, Lu, Mitchell, Batra, Lawrence~Zitnick, and Parikh}]{Antol_2015_ICCV}
Stanislaw Antol, Aishwarya Agrawal, Jiasen Lu, Margaret Mitchell, Dhruv Batra, C~Lawrence~Zitnick, and Devi Parikh. 2015.
\newblock Vqa: Visual question answering.
\newblock In \emph{The IEEE International Conference on Computer Vision (ICCV)}.

\bibitem[{Ba et~al.(2016)Ba, Kiros, and Hinton}]{Ba_2016_arxiv}
Jimmy~Lei Ba, Jamie~Ryan Kiros, and Geoffrey~E Hinton. 2016.
\newblock Layer normalization.
\newblock \emph{arXiv preprint arXiv:1607.06450}.

\bibitem[{Baevski et~al.(2020)Baevski, Zhou, Mohamed, and Auli}]{baevski2020wav2vec}
Alexei Baevski, Yuhao Zhou, Abdelrahman Mohamed, and Michael Auli. 2020.
\newblock wav2vec 2.0: A framework for self-supervised learning of speech representations.
\newblock \emph{Advances in Neural Information Processing Systems}, 33:12449--12460.

\bibitem[{Banerjee and Lavie(2005)}]{banerjee2005meteor}
Satanjeev Banerjee and Alon Lavie. 2005.
\newblock Meteor: An automatic metric for mt evaluation with improved correlation with human judgments.
\newblock In \emph{Proceedings of the acl workshop on intrinsic and extrinsic evaluation measures for machine translation and/or summarization}, pages 65--72.

\bibitem[{Carreira and Zisserman(2017)}]{carreira2017quo}
Joao Carreira and Andrew Zisserman. 2017.
\newblock Quo vadis, action recognition? a new model and the kinetics dataset.
\newblock In \emph{proceedings of the IEEE Conference on Computer Vision and Pattern Recognition}, pages 6299--6308.

\bibitem[{Chu et~al.(2020)Chu, Lin, Hsu, and Ku}]{Chu_2020_DSTC8}
Yun-Wei Chu, Kuan-Yen Lin, Chao-Chun Hsu, and Lun-Wei Ku. 2020.
\newblock Multi-step joint-modality attention network for scene-aware dialogue system.
\newblock In \emph{DSTC8 at AAAI2020 workshop}.

\bibitem[{Devlin et~al.(2018)Devlin, Chang, Lee, and Toutanova}]{devlin2018bert}
Jacob Devlin, Ming-Wei Chang, Kenton Lee, and Kristina Toutanova. 2018.
\newblock Bert: Pre-training of deep bidirectional transformers for language understanding.
\newblock \emph{arXiv preprint arXiv:1810.04805}.

\bibitem[{Garofolo(1993)}]{garofolo1993timit}
John~S Garofolo. 1993.
\newblock Timit acoustic phonetic continuous speech corpus.
\newblock \emph{Linguistic Data Consortium, 1993}.

\bibitem[{Gemmeke et~al.(2017)Gemmeke, Ellis, Freedman, Jansen, Lawrence, Moore, Plakal, and Ritter}]{gemmeke2017audio}
Jort~F Gemmeke, Daniel~PW Ellis, Dylan Freedman, Aren Jansen, Wade Lawrence, R~Channing Moore, Manoj Plakal, and Marvin Ritter. 2017.
\newblock Audio set: An ontology and human-labeled dataset for audio events.
\newblock In \emph{2017 IEEE international conference on acoustics, speech and signal processing (ICASSP)}, pages 776--780. IEEE.

\bibitem[{Geng et~al.(2021)Geng, Gao, Chatterjee, Hori, Le~Roux, Zhang, Li, and Cherian}]{geng2021dynamic}
Shijie Geng, Peng Gao, Moitreya Chatterjee, Chiori Hori, Jonathan Le~Roux, Yongfeng Zhang, Hongsheng Li, and Anoop Cherian. 2021.
\newblock Dynamic graph representation learning for video dialog via multi-modal shuffled transformers.
\newblock In \emph{Proceedings of the AAAI Conference on Artificial Intelligence}, volume~35, pages 1415--1423.

\bibitem[{Hershey et~al.(2017)Hershey, Chaudhuri, Ellis, Gemmeke, Jansen, Moore, Plakal, Platt, Saurous, Seybold et~al.}]{hershey2017cnn}
Shawn Hershey, Sourish Chaudhuri, Daniel~PW Ellis, Jort~F Gemmeke, Aren Jansen, R~Channing Moore, Manoj Plakal, Devin Platt, Rif~A Saurous, Bryan Seybold, et~al. 2017.
\newblock Cnn architectures for large-scale audio classification.
\newblock In \emph{2017 ieee international conference on acoustics, speech and signal processing (icassp)}, pages 131--135. IEEE.

\bibitem[{Hori et~al.(2020)Hori, Cherian, Hori, and Marks}]{hori2020audio}
C~Hori, A~Cherian, T~Hori, and TK~Marks. 2020.
\newblock Audio visual scene-aware dialog (avsd) track for natural language generation in dstc8.
\newblock In \emph{AAAI-DSTC8 Workshop}.

\bibitem[{Hori et~al.(2019{\natexlab{a}})Hori, Alamri, Wang, Wichern, Hori, Cherian, Marks, Cartillier, Lopes, Das, Essa, Batra, and Parikh}]{Hori_2019_ICASSP}
Chiori Hori, Huda Alamri, Jue Wang, Gordon Wichern, Takaaki Hori, Anoop Cherian, Tim~K. Marks, Vincent Cartillier, Raphael~Gontijo Lopes, Abhishek Das, Irfan Essa, Dhruv Batra, and Devi Parikh. 2019{\natexlab{a}}.
\newblock End-to-end audio visual scene-aware dialog using multimodal attention-based video features.
\newblock In \emph{IEEE International Conference on Acoustics, Speech and Signal Processing (ICASSP)}.

\bibitem[{Hori et~al.(2019{\natexlab{b}})Hori, Cherian, Marks, and Hori}]{Hori_2019_Interspeech}
Chiori Hori, Anoop Cherian, Tim~K. Marks, and Takaaki Hori. 2019{\natexlab{b}}.
\newblock Joint student-teacher learning for audio-visual scene-aware dialog.
\newblock In \emph{Proceedings of the Interspeech}.

\bibitem[{Kay et~al.(2017)Kay, Carreira, Simonyan, Zhang, Hillier, Vijayanarasimhan, Viola, Green, Back, Natsev et~al.}]{kay2017kinetics}
Will Kay, Joao Carreira, Karen Simonyan, Brian Zhang, Chloe Hillier, Sudheendra Vijayanarasimhan, Fabio Viola, Tim Green, Trevor Back, Paul Natsev, et~al. 2017.
\newblock The kinetics human action video dataset.
\newblock \emph{arXiv preprint arXiv:1705.06950}.

\bibitem[{Kim et~al.(2021)Kim, Yoon, Kim, and Yoo}]{kim2021structured}
Junyeong Kim, Sunjae Yoon, Dahyun Kim, and Chang~D Yoo. 2021.
\newblock Structured co-reference graph attention for video-grounded dialogue.
\newblock In \emph{Proceedings of the AAAI Conference on Artificial Intelligence}, volume~35, pages 1789--1797.

\bibitem[{Le et~al.(2022)Le, Chen, and Hoi}]{le2022vgnmn}
Hung Le, Nancy Chen, and Steven Hoi. 2022.
\newblock Vgnmn: Video-grounded neural module networks for video-grounded dialogue systems.
\newblock In \emph{Proceedings of the 2022 Conference of the North American Chapter of the Association for Computational Linguistics: Human Language Technologies}, pages 3377--3393.

\bibitem[{Le and Chen(2020)}]{Le_2020_DSTC8}
Hung Le and Nancy~F. Chen. 2020.
\newblock Multimodal transformer with pointer network for the dstc8.
\newblock In \emph{DSTC8 at AAAI2020 workshop}.

\bibitem[{Le et~al.(2021)Le, Chen, and Hoi}]{le2021learning}
Hung Le, Nancy~F Chen, and Steven~CH Hoi. 2021.
\newblock Learning reasoning paths over semantic graphs for video-grounded dialogues.
\newblock \emph{arXiv preprint arXiv:2103.00820}.

\bibitem[{Le et~al.(2019)Le, Sahoo, Chen, and Hoi}]{Le_2019_ACL}
Hung Le, Doyen Sahoo, Nancy Chen, and Steven~C.H. Hoi. 2019.
\newblock Multimodal transformer networks for end-to-end video-grounded dialogue systems.
\newblock In \emph{Proceedings of the 57th Annual Meeting of the Association for Computational Linguistics (ACL)}.

\bibitem[{Lee et~al.(2020)Lee, Yoon, Dernoncourt, Kim, Bui, and Jung}]{Lee_2020_DSTC8}
Hwanhee Lee, Seunghyun Yoon, Franck Dernoncourt, Doo~Soon Kim, Trung Bui, and Kyomin Jung. 2020.
\newblock Dstc8-avsd: Multimodal semantic transformer network with retrieval style word generator.
\newblock In \emph{DSTC8 at AAAI2020 workshop}.

\bibitem[{Li et~al.(2021{\natexlab{a}})Li, Li, Zhang, Feng, and Zhou}]{li2021bridging}
Zekang Li, Zongjia Li, Jinchao Zhang, Yang Feng, and Jie Zhou. 2021{\natexlab{a}}.
\newblock Bridging text and video: A universal multimodal transformer for audio-visual scene-aware dialog.
\newblock \emph{IEEE/ACM Transactions on Audio, Speech, and Language Processing}, 29:2476--2483.

\bibitem[{Li et~al.(2021{\natexlab{b}})Li, Li, Zhang, Feng, and Zhou}]{Li_2021_tip}
Zekang Li, Zongjia Li, Jinchao Zhang, Yang Feng, and Jie Zhou. 2021{\natexlab{b}}.
\newblock Bridging text and video: A universal multimodal transformer for audio-visual scene-aware dialog.
\newblock \emph{IEEE/ACM Transactions on Audio, Speech, and Language Processing}, 29:2476--2483.

\bibitem[{Lin(2004)}]{lin2004rouge}
Chin-Yew Lin. 2004.
\newblock Rouge: A package for automatic evaluation of summaries.
\newblock In \emph{Text summarization branches out}, pages 74--81.

\bibitem[{Lin et~al.(2019)Lin, Hsu, Chen, and Ku}]{Lin_2019_DSTC7}
Kuan-Yen Lin, Chao-Chun Hsu, Yun-Nung Chen, and Lun-Wei Ku. 2019.
\newblock Entropy-enhanced multimodal attention model for scene-aware dialogue generation.
\newblock In \emph{DSTC7 at AAAI2019 workshop}.

\bibitem[{Loshchilov and Hutter(2017)}]{loshchilov2017decoupled}
Ilya Loshchilov and Frank Hutter. 2017.
\newblock Decoupled weight decay regularization.
\newblock \emph{arXiv preprint arXiv:1711.05101}.

\bibitem[{Niu et~al.(2021)Niu, Tang, Zhang, Lu, Hua, and Wen}]{niu2021counterfactual}
Yulei Niu, Kaihua Tang, Hanwang Zhang, Zhiwu Lu, Xian-Sheng Hua, and Ji-Rong Wen. 2021.
\newblock Counterfactual vqa: A cause-effect look at language bias.
\newblock In \emph{Proceedings of the IEEE/CVF Conference on Computer Vision and Pattern Recognition}, pages 12700--12710.

\bibitem[{Panayotov et~al.(2015)Panayotov, Chen, Povey, and Khudanpur}]{panayotov2015librispeech}
Vassil Panayotov, Guoguo Chen, Daniel Povey, and Sanjeev Khudanpur. 2015.
\newblock Librispeech: an asr corpus based on public domain audio books.
\newblock In \emph{2015 IEEE international conference on acoustics, speech and signal processing (ICASSP)}, pages 5206--5210. IEEE.

\bibitem[{Papineni et~al.(2002)Papineni, Roukos, Ward, and Zhu}]{papineni2002bleu}
Kishore Papineni, Salim Roukos, Todd Ward, and Wei-Jing Zhu. 2002.
\newblock Bleu: a method for automatic evaluation of machine translation.
\newblock In \emph{Proceedings of the 40th annual meeting of the Association for Computational Linguistics}, pages 311--318.

\bibitem[{Pham et~al.(2022)Pham, Le, Le, Phuong, and Tran}]{pham2022video}
Hoang-Anh Pham, Thao~Minh Le, Vuong Le, Tu~Minh Phuong, and Truyen Tran. 2022.
\newblock Video dialog as conversation about objects living in space-time.
\newblock In \emph{European Conference on Computer Vision}, pages 710--726. Springer.

\bibitem[{Radford et~al.(2018)Radford, Narasimhan, Salimans, Sutskever et~al.}]{radford2018improving}
Alec Radford, Karthik Narasimhan, Tim Salimans, Ilya Sutskever, et~al. 2018.
\newblock Improving language understanding by generative pre-training.

\bibitem[{Radford et~al.(2019)Radford, Wu, Child, Luan, Amodei, Sutskever et~al.}]{radford2019language}
Alec Radford, Jeffrey Wu, Rewon Child, David Luan, Dario Amodei, Ilya Sutskever, et~al. 2019.
\newblock Language models are unsupervised multitask learners.
\newblock \emph{OpenAI blog}, 1(8):9.

\bibitem[{Raffel et~al.(2020)Raffel, Shazeer, Roberts, Lee, Narang, Matena, Zhou, Li, Liu et~al.}]{raffel2020exploring}
Colin Raffel, Noam Shazeer, Adam Roberts, Katherine Lee, Sharan Narang, Michael Matena, Yanqi Zhou, Wei Li, Peter~J Liu, et~al. 2020.
\newblock Exploring the limits of transfer learning with a unified text-to-text transformer.
\newblock \emph{J. Mach. Learn. Res.}, 21(140):1--67.

\bibitem[{Sanabria et~al.(2019)Sanabria, Palaskar, and Metze}]{Sanabria_2019_DSTC7}
Ramon Sanabria, Shruti Palaskar, and Florian Metze. 2019.
\newblock Cmu sinbad's submissino for the dstc7 avsd challenge.
\newblock In \emph{DSTC7 at AAAI2019 workshop}.

\bibitem[{Sigurdsson et~al.(2016)Sigurdsson, Varol, Wang, Farhadi, Laptev, and Gupta}]{sigurdsson2016hollywood}
Gunnar~A Sigurdsson, G{\"u}l Varol, Xiaolong Wang, Ali Farhadi, Ivan Laptev, and Abhinav Gupta. 2016.
\newblock Hollywood in homes: Crowdsourcing data collection for activity understanding.
\newblock In \emph{European Conference on Computer Vision}, pages 510--526. Springer.

\bibitem[{Vaswani et~al.(2017)Vaswani, Shazeer, Parmar, Uszkoreit, Jones, Gomez, Kaiser, and Polosukhin}]{vaswani2017attention}
Ashish Vaswani, Noam Shazeer, Niki Parmar, Jakob Uszkoreit, Llion Jones, Aidan~N Gomez, {\L}ukasz Kaiser, and Illia Polosukhin. 2017.
\newblock Attention is all you need.
\newblock \emph{Advances in neural information processing systems}, 30.

\bibitem[{Vedantam et~al.(2015)Vedantam, Lawrence~Zitnick, and Parikh}]{vedantam2015cider}
Ramakrishna Vedantam, C~Lawrence~Zitnick, and Devi Parikh. 2015.
\newblock Cider: Consensus-based image description evaluation.
\newblock In \emph{Proceedings of the IEEE conference on computer vision and pattern recognition}, pages 4566--4575.

\bibitem[{Wang et~al.(2022)Wang, Bao, Dong, Bjorck, Peng, Liu, Aggarwal, Mohammed, Singhal, Som et~al.}]{wang2022image}
Wenhui Wang, Hangbo Bao, Li~Dong, Johan Bjorck, Zhiliang Peng, Qiang Liu, Kriti Aggarwal, Owais~Khan Mohammed, Saksham Singhal, Subhojit Som, et~al. 2022.
\newblock Image as a foreign language: Beit pretraining for all vision and vision-language tasks.
\newblock \emph{arXiv preprint arXiv:2208.10442}.

\bibitem[{Wu et~al.(2016)Wu, Schuster, Chen, Le, Norouzi, Macherey, Krikun, Cao, Gao, Macherey et~al.}]{wu2016google}
Yonghui Wu, Mike Schuster, Zhifeng Chen, Quoc~V Le, Mohammad Norouzi, Wolfgang Macherey, Maxim Krikun, Yuan Cao, Qin Gao, Klaus Macherey, et~al. 2016.
\newblock Google's neural machine translation system: Bridging the gap between human and machine translation.
\newblock \emph{arXiv preprint arXiv:1609.08144}.

\bibitem[{Xie and Iacobacci(2020)}]{Xie_2020_DSTC8}
Huiyuan Xie and Ignacio Iacobacci. 2020.
\newblock Audio visual scene-aware dialog system using dynamic memory networks.
\newblock In \emph{DSTC8 at AAAI2020 workshop}.

\bibitem[{Yoon et~al.(2022{\natexlab{a}})Yoon, Yoon, Harvill, Yoon, Hasegawa-Johnson, and Yoo}]{yoon2022smsmix}
Hee~Suk Yoon, Eunseop Yoon, John Harvill, Sunjae Yoon, Mark Hasegawa-Johnson, and Chang~D Yoo. 2022{\natexlab{a}}.
\newblock Smsmix: Sense-maintained sentence mixup for word sense disambiguation.
\newblock \emph{arXiv preprint arXiv:2212.07072}.

\bibitem[{Yoon et~al.(2022{\natexlab{b}})Yoon, Hong, Yoon, Kim, Kim, Yoon, and Yoo}]{yoon2022selective}
Sunjae Yoon, Ji~Woo Hong, Eunseop Yoon, Dahyun Kim, Junyeong Kim, Hee~Suk Yoon, and Chang~D Yoo. 2022{\natexlab{b}}.
\newblock Selective query-guided debiasing for video corpus moment retrieval.
\newblock In \emph{European Conference on Computer Vision}, pages 185--200. Springer.

\bibitem[{Yoon et~al.(2023)Yoon, Koo, Kim, and Yoo}]{yoon2023scanet}
Sunjae Yoon, Gwanhyeong Koo, Dahyun Kim, and Chang~D Yoo. 2023.
\newblock Scanet: Scene complexity aware network for weakly-supervised video moment retrieval.
\newblock In \emph{Proceedings of the IEEE/CVF International Conference on Computer Vision}, pages 13576--13586.

\bibitem[{Yoon et~al.(2022{\natexlab{c}})Yoon, Yoon, Yoon, Kim, and Yoo}]{yoon-etal-2022-information}
Sunjae Yoon, Eunseop Yoon, Hee~Suk Yoon, Junyeong Kim, and Chang Yoo. 2022{\natexlab{c}}.
\newblock \href {https://doi.org/10.18653/v1/2022.emnlp-main.280} {Information-theoretic text hallucination reduction for video-grounded dialogue}.
\newblock In \emph{Proceedings of the 2022 Conference on Empirical Methods in Natural Language Processing}, pages 4182--4193, Abu Dhabi, United Arab Emirates. Association for Computational Linguistics.

\bibitem[{Zheng et~al.(2022)Zheng, Huang, Chen, Peng, and Liu}]{zheng2022weakly}
Minghang Zheng, Yanjie Huang, Qingchao Chen, Yuxin Peng, and Yang Liu. 2022.
\newblock Weakly supervised temporal sentence grounding with gaussian-based contrastive proposal learning.
\newblock In \emph{Proceedings of the IEEE/CVF Conference on Computer Vision and Pattern Recognition}, pages 15555--15564.

\end{thebibliography}
\bibliographystyle{acl_natbib}

\appendix

\clearpage
\appendix
\label{sec:appendix}
\section{Experimental Details}
\paragraph{Training.} Proposed HEAR is trained on NVIDIA Quadro RTX 8000 (48GB of memory) GPU. 
The optimization details are as follows.
The AdamW optimizer \cite{loshchilov2017decoupled} is used with parameters as $\beta_{1}$ = 0.9, $\beta_{2}$ = 0.999, and $\epsilon$ = 10e-8. 
Due to the efficient usage of resources, the previous 3 question-answer pairs (\ie $\{C,(Q^{r-3},A^{r-3}),(Q^{r-2},A^{r-2}),(Q^{r-1},A^{r-1})\}$) as a dialogue history are introduced into HEAR for answering current questions $Q^{r}$.
For the learning rate, we first set it as $lr = 6.24e-5$ and it linearly decreases following a piecewise linear curve up to $lr = 3.63e-10$ and the model is trained the model during 15 epochs. 
The hyperparameters are $\delta = 0.05$ for the margin in $\mathcal{L}_{RLE}$.
For the joint $d$-dimensional space, all modalities (\ie video, audio, words) are embedded  $d=768$ dimensional space.
The best model is decided by the lowest loss of the validation set on AVSD@DSTC7 (\ie AVSD@DSTC8 contains only test set for evaluation.).
It takes 14 hours to finish the training and the model is fully optimized in about 10 hours.
\paragraph{Inference.} In the inference, answer word tokens are generated in a sequence with a probability, where the beam search is applied to avoid prompt word selection with a beam size of 5.
The max length for the answer word tokens is set to 20
with a length penalty of 0.3.
In the Experiment of the main paper, every performance of HEAR is averaged 15 times with a random seed number.
\section{Additional Experiments}
To improve the reproducibility of proposed modules (\ie SAL, RLE) in HEAR, we give detailed explanations about them with additional results and illustrations of the method. 
\subsection{Keyword lists for keyword-based audio sensing}
\label{b.1}
\begin{figure}[t!]
	\centering
        \includegraphics[width=\linewidth]{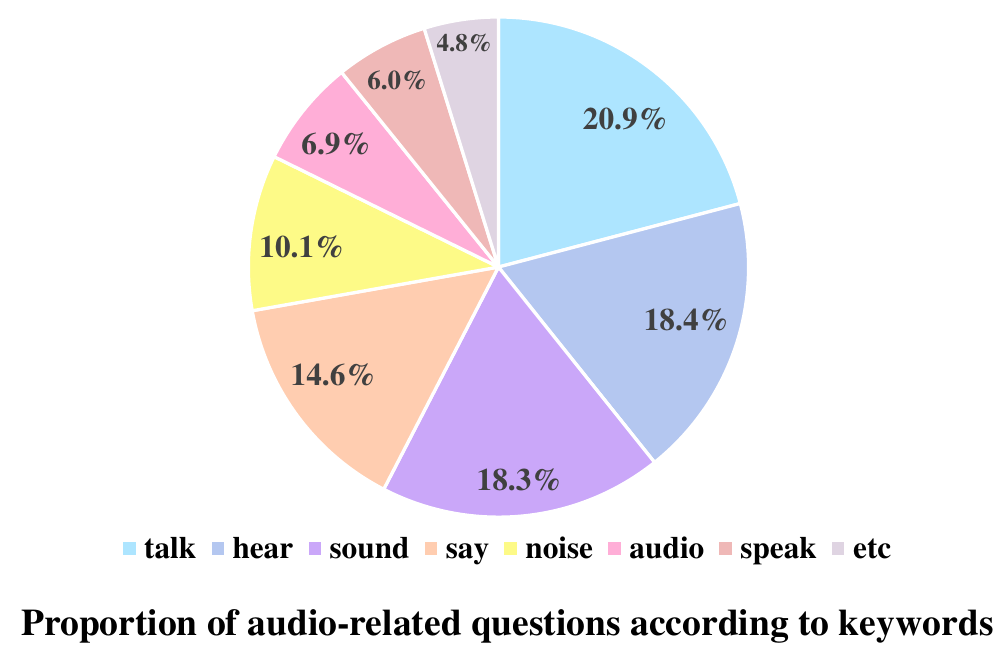}
	\caption{Chart for representing the proportion of audio-related questions according to keywords in $\text{W}_{key}$.}
	\label{fig:app0}
\end{figure}

For details of Section \ref{sec:4.3}, we list the keywords that we used for keyword-based audio sensing as given below:
\begin{equation}
\begin{aligned}
\text{W}_{key} = \{&\text{noise}, \text{sound}, \text{voice}, \text{speech},\\ &\text{speak}, \text{talk}, \text{listen}, \text{hear}, \text{say}, \text{sing},\\ &\text{music}, \text{audio}, \text{call}, \text{hum}, \text{loud},\\ &\text{tones}, \text{utter}, \text{volume}, \text{song}\},
\end{aligned}
\end{equation}
where we also considered all the keywords' plural forms in $\text{W}_{key}$.
Figure \ref{fig:app0} summarizes the proportion of audio-related questions according to each keyword, where `talk', `hear', and `sound' are the most related words to the audio questions.
We additionally considered more keywords (\eg `tell', `background'), but they rather degrade the motivation of keyword-based audio sensing by finding other questions (\eg `can you tell me...', 'can you see something in the background') excluding audio-related questions.
\begin{table}
  \centering
  \small
  \begin{tabular}{lccc}
  \toprule
   Number of keywords & BLEU 1 & ROUGE-L & CIDEr\\     \midrule
   17 & 0.327 & 0.367 & 1.458\\ 
   18 & 0.333 & 0.368 & 1.466\\ 
   19 & 0.335 & 0.371 & 1.469\\ 
   20 & 0.331 & 0.366 & 1.468\\ 
   21 & 0.324 & 0.363 & 1.465\\ 
   22 & 0.321 & 0.361 & 1.460\\ \bottomrule
  \end{tabular}
  \caption{Performance variations according to the number of keywords for Keyword-based Audio Sensing on AVSD@DSTC7 validation set}
  \label{tab:audio_que}
\end{table}
To be specific, we extend the evaluations for SAL in terms of Keyword-based Audio Sensing (Table A).
We investigate the performance variations according to the quantity of audio-related keywords utilized. The results show a positive correlation between the number of keywords and the enhancement. 
However, surpassing a specific number (\ie 19) leads to a reversal in this trend, revealing a detrimental correlation. 
This implies that the pool of keywords starts to include words that lack sensibility in effectively categorizing audio-related questions. Thus, we employ 19 keywords for SAL listed in Appendix \ref{b.1}
%
\begin{figure*}[t]
	\centering
	\includegraphics[width=\linewidth]{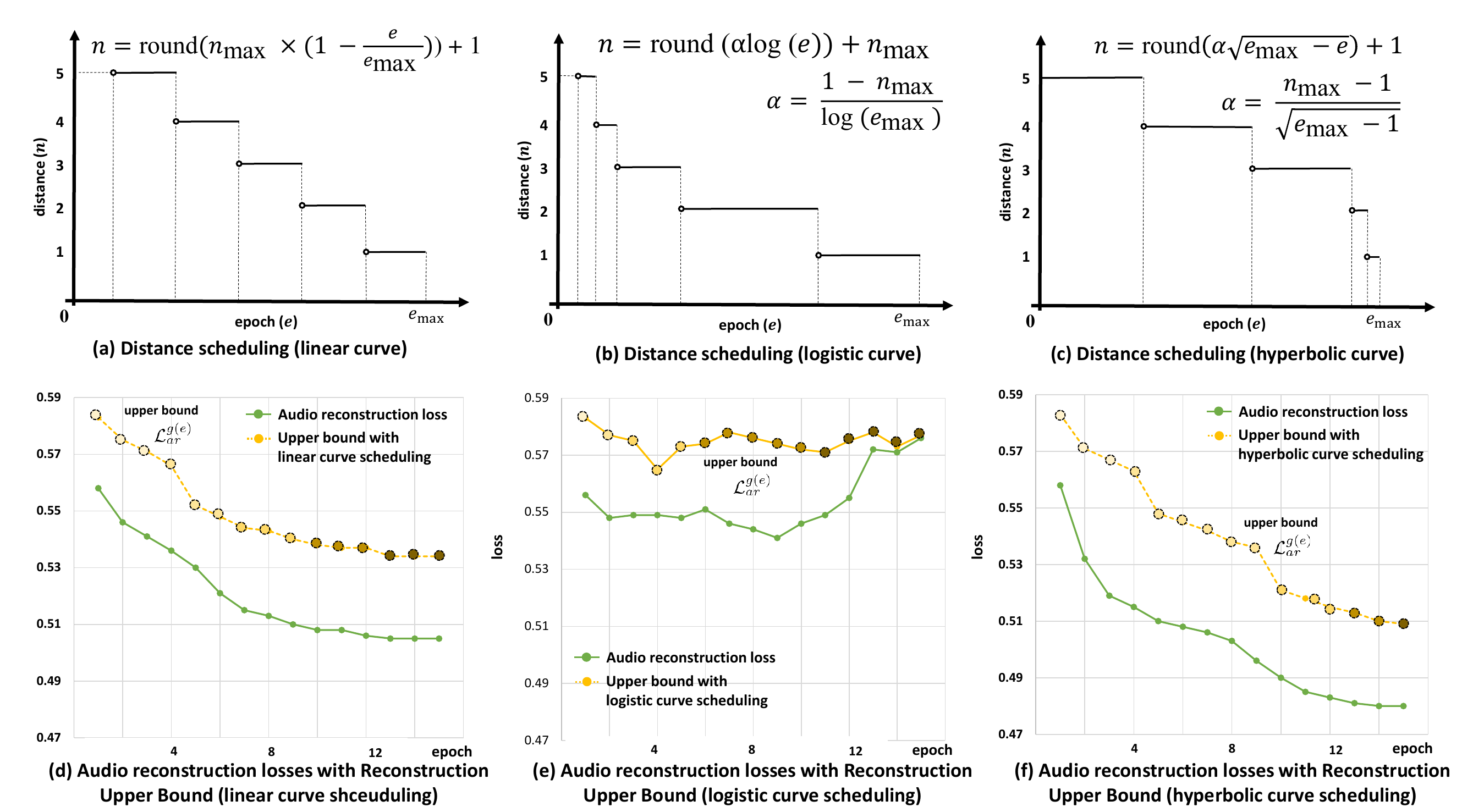}
	\caption{Illustrations of distance scheduling for surrounding masking. The distance $n$ decides how much mask video and audio features for establishing upper bound $\mathcal{L}_{ar}^{n}$: (a) linear curve, (b) logistic curve, (c) hyperbolic curve. Audio reconstruction losses with reconstruction upper bound using (d) linear curve scheduling, (e) logistic curve scheduling, (f) hyperbolic curve scheduling.}
	\label{fig:app1}
\end{figure*}
\subsection{Surrounding Masking Scheduling}
For the reconstruction upper bound in Section \ref{sec:4.4}, we provide further ablation studies about distance scheduling for surrounding masking.
The surrounding mask is designed for promoting ranking loss in $\mathcal{L}_{rub}$ by ensuring that the audio reconstruction $\mathcal{L}_{ar}$ based on the surrounding information has an improved quality than audio reconstruction $\mathcal{L}_{ar}^{n}$ without the surrounding information.
%
%
To make more effective $\mathcal{L}_{rub}$, $\mathcal{L}_{ar}^{n}$ can be designed to minimize its loss by By reducing the distance $n$ of the surrounding mask.
To this end, we leverage the extent of the surrounding mask under our designed various modelings $n = g(e)$ for masking scheduling in Figure \ref{fig:app1}.
We narrowed down the extent of surrounding masking based on three different curves: linear curve, logistic curve, and hyperbolic curve.
The hyperbolic curve and linear curve show effectiveness in optimizing validation loss.
However, the logistic curve shows some deterioration in the optimization.
We think this is because the logistic curve makes the surrounding masking applied to a very narrow area from the beginning of training, which acts like a hard negative that is almost similar to the positive reconstruction (\ie reconstruction from only masked target audio).
Providing hard negatives in early training is considered to hinder optimization when the learnable weights of the model are not properly trained.
Therefore, surrounding masking with hyperbolic scheduling is the most effective.
%
%
%
\begin{table}
  \centering
  \begin{tabular}{cccc}
  \Xhline{3\arrayrulewidth}
    \multicolumn{1}{c}{$n_{\textrm{max}}$} & \multicolumn{1}{c}{\multirow{1}{*}{BLEU1}} & \multicolumn{1}{c}{\multirow{1}{*}{ROUGE-L}} & \multicolumn{1}{c}{\multirow{1}{*}{CIDEr}} \\ 
    \Xhline{2\arrayrulewidth}
       2 &0.310  &0.362 &1.436\\
       3 &\bf{0.348}  &0.371 &1.475\\
       4  &0.341  &\bf{0.384} &\bf{1.492}\\
       5   &0.316  &0.369 &1.438\\
      6  &0.314  &0.368 &1.436\\ \Xhline{3\arrayrulewidth}
  \end{tabular}
  \caption{Ablation study on the distance of negative masking of HEAR on valid split of AVSD@DSTC7.}
  \label{tab:apdx_1}
\end{table}

Table \ref{tab:apdx_1} summarizes the results of HEAR according to the distance $n$ on the validation split of AVSD@DSTC7 under distance scheduling with the hyperbolic curve.
%
%
%
%
When $n_{\textrm{max}}$ is small (\eg $n_{\textrm{max}}$ = 2 or 3), $\mathcal{L}_{ar}^{*}$ makes audio reconstruction based on the nearest video and audio from the target audio.
This can be effective in terms of improving the connectivity of neighboring video and audio at a narrow distance, however since the distance is too close, there is no significant difference from the positive audio reconstruction, which hinders reconstruction.
Based on results in Table \ref{tab:apdx_1}, negative masking was effective when $n_{\textrm{max}}$ = 4 or 5.
In this case, the surrounding modalities' features for audio reconstruction were properly removed by masking, in the meanwhile the surroundings were not masked excessively.
Thus the masking contributes on hard negative audio reconstruction.
When $n_{\textrm{max}}$ is large (\eg $n_{\textrm{max}}$ = 6), the performance again degrades.
For this case we guess that the negative reconstruction is quite different from the positive, so it would not be an effective negative.
These studies were also conducted for a larger $n_{\textrm{max}}$, but no further improvement was confirmed.
\begin{table}
  \centering
  \begin{tabular}{lcccc}
  \Xhline{3\arrayrulewidth}
    \multicolumn{1}{l}{Type} & \multicolumn{1}{c}{\multirow{1}{*}{HEAR}} & \multicolumn{1}{c}{\multirow{1}{*}{THAM}} & \multicolumn{1}{c}{\multirow{1}{*}{RLM}} & \multicolumn{1}{c}{\multirow{1}{*}{GT}} \\ 
    \Xhline{2\arrayrulewidth}
       semantic &2.41  &2.15 &1.81 &2.86\\
       grammatical &2.35  &2.24 &2.16 &2.76\\
       fluency  &2.62  &2.58 &2.41 &2.81\\ \Xhline{3\arrayrulewidth}
  \end{tabular}
  \caption{Human evaluation on responses to 50 audio-related questions with respect to semantic adequacy, grammatical correctness, fluency. The scores are assigned as (``$1$: not at all", ``$2$: neutral", ``$3$: correct"), GT: Ground-Truth response.}
  \label{tab:human_eval}
\end{table}
\section{Human Evaluation}
To assess the improvement of HEAR framework in the responses to audio-related questions, we conduct a Human Evaluation.
We selected a total of 50 audio-related questions and generate their responses from 3 models (\ie HEAR, THAM, RLM).
We evaluate all responses based on a scale of 1 to 3, considering three key perspectives: semantic adequacy, grammatical correctness, and fluency.
Each score denotes that ``$1$: not at all", ``$2$: neutral", ``$3$: correct".
Table \ref{tab:human_eval} shows the human evaluation scores based on 11 evaluators.
Evaluators rate the responses based on three categories, where HEAR outperformed recent runner models in all three categories, receiving higher ratings.
Furthermore, through empirical validation, we have confirmed the presence of inappropriate answer responses within the Ground-Truth. 
As a result, the human evaluation scores appropriately reflect this by not consistently obtaining a score of 3.0 or a value close to it.
\begin{figure}[th]
	\centering
	\includegraphics[width=\linewidth]{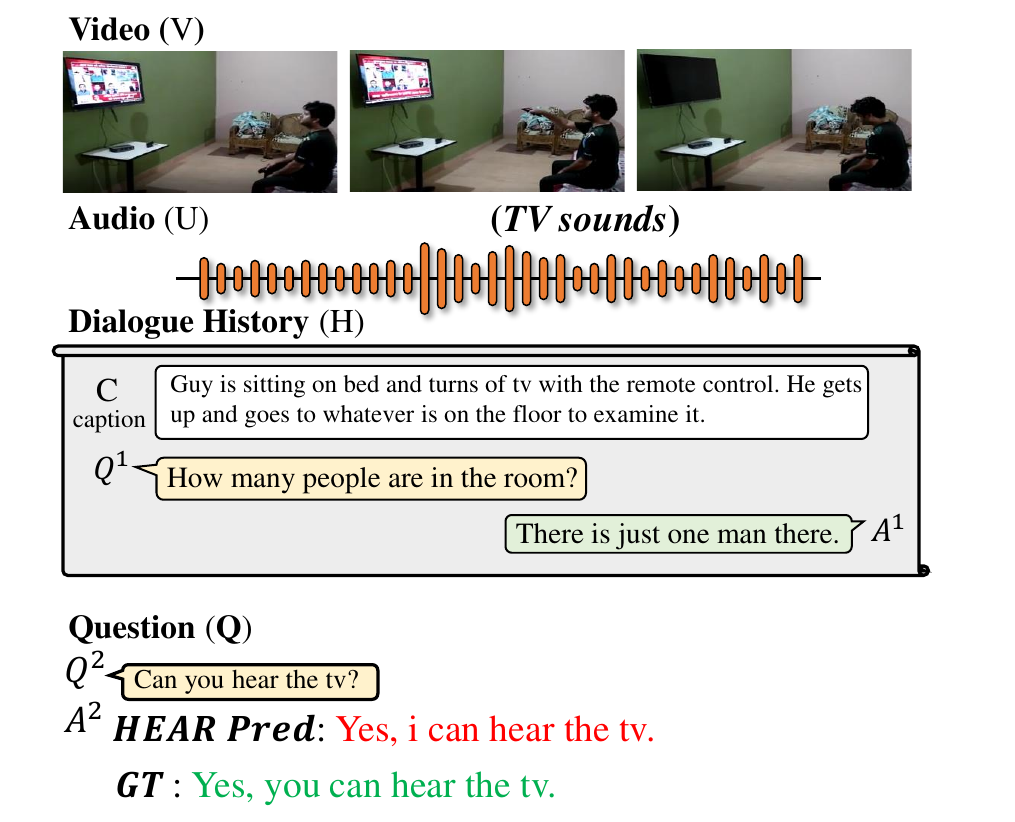}
	\caption{Illustration of additional results of HEAR}
	\label{fig:add_result}
\end{figure}
\begin{figure}[th]
	\centering
	\includegraphics[width=\linewidth]{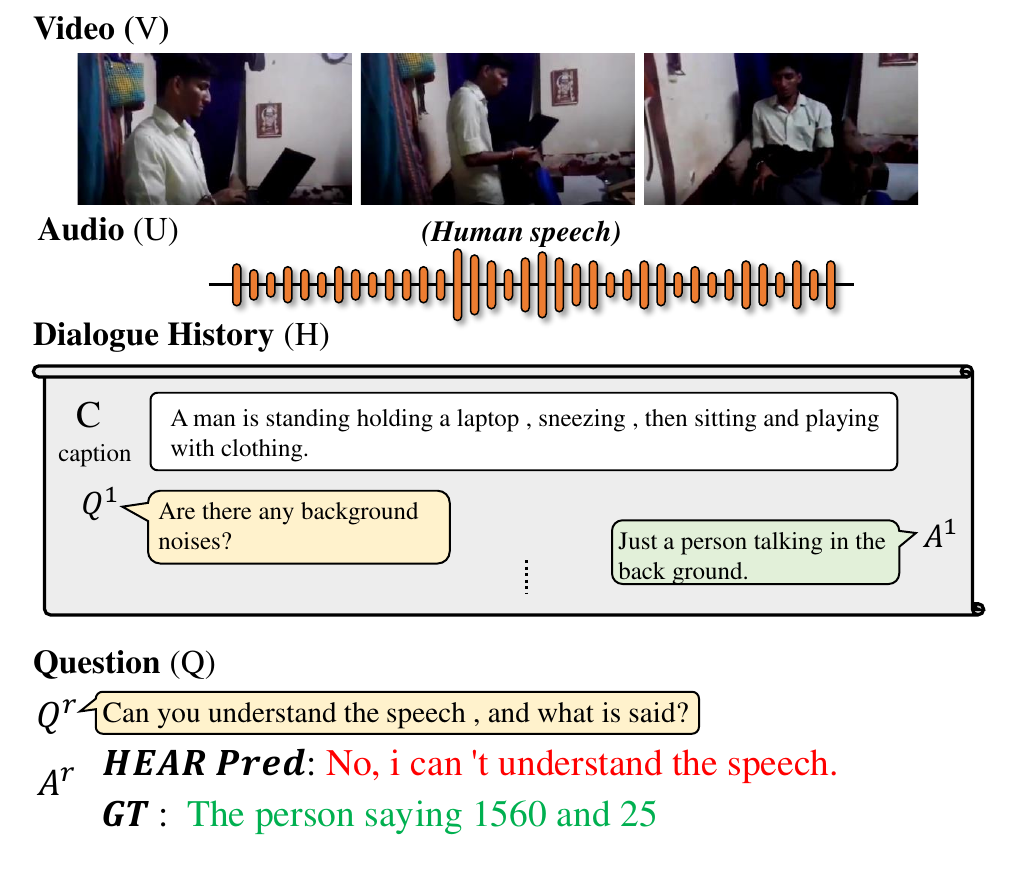}
	\caption{Illustration of failure case of HEAR}
	\label{fig:failure}
\end{figure}
\section{Additional results and Failure case}
Figure \ref{fig:add_result} illustrates additional results of HEAR framework.
When presented with the question, "Can you hear the TV?" our HEAR provides an affirmative response, stating, "Yes, I can hear the TV." 
This response is generated by considering both the audio input of the TV and the visual context of the TV image.
Although our proposed HEAR improves the understanding of audio, there were instances of failure when it comes to questions pertaining to speech recognition.
As shown in Figure \ref{fig:failure}, the HEAR framework demonstrates limitations in accurately comprehending the language within audio inputs. 
Our empirical studies found that this challenge is prevalent among various dialogue language models, including our model.
It is noted that the current audio feature used in the HEAR framework is pre-trained with the environmental sounds dataset, such as AudioSet \cite{gemmeke2017audio}. 
However, it appears that further data training is necessary to enhance the system's ability to understand speech effectively.
To address this challenge, our future work will focus on incorporating audio features, such as wav2vec 2.0 \cite{baevski2020wav2vec}, which have been trained specifically for audio speech recognition tasks. By leveraging these advanced audio features, we aim to enhance the model's ability to accurately understand and process speech-related information.

\end{document}